\documentclass[lettersize,journal]{IEEEtran}
\usepackage{amsmath,amsfonts}
\usepackage{algorithmic}
\usepackage{balance}
\usepackage{algorithm}
\usepackage{array}
\usepackage{textcomp}
\usepackage{stfloats}
\usepackage{url}
\usepackage{verbatim}
\usepackage{graphicx}
\usepackage{cite}
\usepackage{multirow}
\usepackage{makecell}
\usepackage{tabularx}
\usepackage{booktabs}
\usepackage{longtable}
\usepackage{subcaption} 
\usepackage{ragged2e}
\usepackage{multicol}
\usepackage{hyperref}
\usepackage{booktabs}
\usepackage{siunitx}
\usepackage[justification=centering]{caption}
\begin{document}

\title{SenseRay-3D: Generalizable and Physics-Informed Framework for End-to-End Indoor Propagation Modeling}

\author{Yu Zheng, Kezhi Wang,~\IEEEmembership{Senior Member, IEEE}, Wenji Xi, Gang Yu,

Jiming Chen, Jie Zhang,~\IEEEmembership{Senior Member, IEEE}
\thanks{Yu Zheng, Wenji Xi, and Gang Yu are with the School of Electrical and Electronic Engineering, The University of Sheffield, S10 2TN Sheffield, U.K. (e-mail: \{yzheng120, wxi4\}@sheffield.ac.uk;gangyu@ieee.org).}
\thanks{Kezhi Wang is with the Department of Computer Science, Brunel University London, UB8 3PH, U.K. (e-mail: kezhi.wang@brunel.ac.uk).}

\thanks{Jiming Chen is with Ranplan Wireless Network Design Ltd., Cambridge
CB23 3UY, U.K. (e-mail: jiming.chen@ranplanwireless.com).}
\thanks{Jie Zhang is with the R\&D Department, Cambridge AI+ Ltd., CB23 3UY
Cambridge, U.K., and also with the R\&D Department, Ranplan Wireless
Network Design Ltd., CB23 3UY Cambridge, U.K. (e-mail: Jie.Zhang@ranplanwireless.com).}
\thanks{This work is supported in part by Eureka COMET/UKRI Innovate UK project, Grant No. 10099265 and Horizon Europe COVER project, No. 101086228 (with funding from UKRI grant EP/Y028031/1). K. Wang would like to acknowledge support from the Royal Society Industry Fellowship (IF/R2/23200104).
}
}

\markboth{Journal of \LaTeX\ Class Files,~Vol.~14, No.~8, August~2021}%
{Shell \MakeLowercase{\textit{et al.}}: A Sample Article Using IEEEtran.cls for IEEE Journals}


\maketitle
\begin{abstract}
Modeling indoor radio propagation is crucial for wireless network planning and optimization. However, existing approaches often rely on labor-intensive manual modeling of geometry and material properties, resulting in limited scalability and efficiency. To overcome these challenges, this paper presents SenseRay-3D, a generalizable and physics-informed end-to-end framework that predicts three-dimensional (3D) path-loss heatmaps directly from RGB-D scans, thereby eliminating the need for explicit geometry reconstruction or material annotation. The proposed framework builds a sensing-driven voxelized scene representation that jointly encodes occupancy, electromagnetic material characteristics, and transmitter–receiver geometry, which is processed by a SwinUNETR-based neural network to infer environmental path-loss relative to free-space path-loss. A comprehensive synthetic indoor propagation dataset is further developed to validate the framework and to serve as a standardized benchmark for future research. Experimental results show that SenseRay-3D achieves a mean absolute error of 4.27 dB on unseen environments and supports real-time inference at 217 ms per sample, demonstrating its scalability, efficiency, and physical consistency. SenseRay-3D paves a new path for sense-driven, generalizable, and physics-consistent modeling of indoor propagation, marking a major leap beyond our pioneering EM DeepRay framework.

\end{abstract}


\section{Introduction}

The rapid growth of wireless communication has placed unprecedented demands on network capacity, coverage, and reliability. More than 80\% of all traffic occurs indoors \cite{9040264}, where users expect seamless connectivity for applications such as video streaming, virtual reality, and internet-of-things. Meeting these expectations requires accurate modeling of indoor radio propagation, which underpins essential network planning tasks such as access point deployment and channel assignment \cite{siomina2007radio}. Reliable propagation modeling also enables operators to evaluate coverage, identify dead zones, and optimize deployment strategies proactively before undertaking costly real-world trials.

However, indoor propagation modeling remains challenging due to the complex nature of indoor environments. Diverse obstacles, irregular layouts, and heterogeneous construction materials give rise to multipath reflections, frequent non-line-of-sight (NLoS) conditions, and material-dependent attenuation \cite{wang2022reflection}. Generally, indoor propagation modeling can be categorized into empirical and deterministic approaches \cite{obeidat2020indoor}. Empirical modeling, such as the COST-231 model \cite{mogensen1999cost} and the ITU-R P.1238 model \cite{ITU-RP1238-12}, provides parametric equations characterized by statistical analysis of measured data \cite{6927716}. Although they are practical and computationally efficient in use, these models generalize poorly to unseen layouts and often yield noticeable deviations from reality \cite{1543252}. In contrast, deterministic approaches, such as ray tracing \cite{103807}, ray launching \cite{10464960}, and full-wave solvers (e.g., the finite-difference time-domain method) \cite{5979136}, explicitly compute the electromagnetic field for every location by solving Maxwell's equations under specific boundary conditions within a specified physical geometry \cite{10599118}. These methods provide high accuracy but incur prohibitive computational costs, particularly in large and richly furnished indoor environments. Moreover, deterministic approaches require high-fidelity environmental representations \cite{10188258}, in which both the detailed reconstruction of indoor geometries and the accurate characterization of material electromagnetic properties are indispensable. These strict requirements substantially increase the preparation overhead and impose considerable demands in terms of manual labor and domain expertise. Such limitations motivate the exploration of machine-learning-based alternatives.

Driven by the rapid progress of machine learning (ML), the ability to efficiently process large-scale data and distill actionable insights has the potential to fundamentally reshape wireless network operation \cite{8743390}. ML-based propagation models promise to balance accuracy and efficiency, providing an attractive alternative to purely empirical or deterministic methods \cite{9496115}. Building on this trend, recent studies have demonstrated the effectiveness of neural networks for propagation modeling tasks \cite{10640063}. For example, multilayer perceptrons have been used to directly learn the mapping between transmitter–receiver coordinates and received signal strength or path-loss \cite{1696368,6748900}. To better exploit spatial correlations, convolutional neural networks have been employed to predict coverage or path-loss from location grids \cite{9354618,9670666}. 

More advanced U-Net variants have been explored for dense path-loss heatmap regression, leveraging their encoder–decoder design with skip connections to preserve fine-grained spatial details \cite{9771088}, and notably, Dr. Jiming Chen and Professor Jie Zhang, contributors of this work, were also co-authors of the EM DeepRay model in \cite{9771088}, the first DNN-based indoor propagation model with generalizable performance and was selected in the special issue to commemorating the 75th anniversary of IEEE Antenna and Propagation Society. Nevertheless, most of these ML-based approaches still rely on detailed environmental modeling to generate training data or input features, which often entails the same time-consuming and labor-intensive processes as deterministic methods. Moreover, the majority of existing works are limited to two-dimensional representations of the environment, without explicitly incorporating full three-dimensional (3D) scene geometry and material information, which limits their generalization to realistic indoor environments.

Meanwhile, advances in sensing and perception technologies have opened new opportunities to enrich propagation modeling. The emerging paradigm of joint communication and sensing, together with 3D perception techniques such as RGB-D scanning, LiDAR, and point cloud reconstruction, provides a natural way to capture both the geometry and the material properties of indoor environments \cite{10430216,10812728}. 

Motivated by this trend, we propose SenseRay-3D, a sensing-driven framework for indoor propagation modeling. It leverages RGB-D data to construct dense, semantically annotated point clouds. These are voxelized into structured feature tensors that jointly encode occupancy, material-aware electromagnetic properties, and distance information. Based on this representation, we utilize a SwinUNETR-based learning model that directly predicts the path-loss at multiple receiver heights, thereby capturing vertical propagation effects such as ceiling and floor reflections, as well as the impact of furniture with varying heights. By integrating sensing, voxel-based representation, and SwinUNETR-based learning, SenseRay-3D provides an efficient and accurate alternative to conventional methods. The main contributions of this paper are summarized:
\begin{enumerate}
    \item We propose SenseRay-3D, an end-to-end, generalizable, and physics-informed framework for indoor propagation modeling that directly predicts 3D path-loss heatmaps from RGB-D scans. The framework integrates geometric and electromagnetic priors into both the data representation and the learning objectives, allowing the network to learn propagation behavior consistent with physical laws. This design removes the need for labor-intensive manual geometry reconstruction and material annotation while achieving strong generalization across diverse indoor environments.
    \item SenseRay-3D employs a SwinUNETR-based  \cite{he2023swinunetrv2} neural predictor for multi-height path-loss prediction, which takes voxelized scene representations encoding occupancy, material properties, and transmitter–receiver geometry as input. The network predicts environmental path-loss relative to the free-space path-loss (FSPL), and the final path-loss heatmaps are reconstructed by adding the predicted residuals to the FSPL baseline at multiple receiver heights. This residual-based design enforces physical consistency, effectively captures both local and global spatial dependencies, and enables efficient modeling of indoor propagation characteristics.
    \item We construct a comprehensive synthetic dataset for validation and future research benchmarking by combining 3D-FRONT layouts \cite{fu20213d}, BlenderProc rendering \cite{Denninger2023}, and Sionna electromagnetic simulations\cite{sionna}. The pipeline generates paired voxel features and ground-truth path-loss heatmap with geometry, material, and multipath effects. This dataset is used to rigorously validate the SenseRay-3D framework and serves as a standardized benchmark for learning-based indoor propagation modeling.
    \item Extensive experiments demonstrate that SenseRay-3D reconstructs indoor path-loss distributions with a mean absolute error of 4.27 dB and supports real-time inference at 217 ms per sample. It confirms the framework’s effectiveness, computational efficiency, and scalability for practical indoor propagation modeling.
\end{enumerate}

The remainder of the article is organized as follows. Section~II provides a brief overview of some existing studies on sensing-driven propagation modeling. Section~III introduces the indoor system model and formulates the path-loss prediction problem. Section~IV presents the proposed SenseRay-3D framework, including the sensing-driven voxelized scene representation, the SwinUNETR-based neural predictor, and the path-loss reconstruction. Section~V outlines the dataset generation pipeline. Section~VI details the simulation setup and presents the experimental evaluation and analysis. Finally, Section~VII concludes the paper by summarizing the main contributions.

\section{Related Work}
\label{sec:related}

Recent advances in joint communication and sensing have enabled the integration of 3D perception techniques such as LiDAR, RGB-D scanning, and photogrammetry into radio propagation analysis \cite{10949588}. By capturing dense geometric and material information of the environment, these sensing modalities offer a new perspective for efficient propagation modeling beyond traditional empirical and deterministic methods.

Early sensing-aided studies primarily relied on LiDAR- or photogrammetry-based geometry reconstruction to support deterministic simulations. 
For instance, Okamura \textit{et al.} proposed an automatic indoor reconstruction method using LiDAR-generated 3D point clouds for ray-tracing simulation in \cite{9977926}. In their approach, ceiling, floor, walls, and indoor objects were segmented from point clouds through height-based histogram analysis and contour detection. The reconstructed polygonal model was then used as input to a deterministic ray-tracing simulator. Similarly, Suga \textit{et al.} proposed an RGB-D sensor–aided indoor radio-map construction framework in \cite{10064304,10949606}, which reconstructs the 3D geometry of walls, ceilings, and furniture from RGB-D images via semantic segmentation and cuboid approximation. The resulting geometric primitives are automatically imported into a ray-tracing simulator to compute the received power distribution. Although these sensing-assisted frameworks reduce the need for manual geometry construction, their dependence on deterministic modeling still entails high computational cost and limited scalability, motivating the exploration of data-driven alternatives that can learn propagation characteristics directly from sensed 3D environments.

To this end, Clark \textit{et al.} introduced PropEM-L \cite{clark2022propem}, a perception-driven framework for radio-signal prediction in multi-robot exploration scenarios. PropEM-L leverages LiDAR-based occupancy grids to encode line-of-sight visibility, shadowing, reflection, and diffraction features, which are then combined with neural networks for received signal strength prediction. This hybrid modeling strategy improves prediction accuracy while retaining physical interpretability, yet it still relies on sparse geometric features and lacks material-awareness, thereby limiting its applicability in complex indoor environments. A more recent work, NeRA \cite{10757803}, proposed a deep-learning-based representation called Neural Reflectance and Attenuation Fields, which jointly models RF attenuation and surface reflectance from voxelized point clouds. Unlike conventional ray-based approaches, NeRA learns both attenuation and reflection behaviors through multiple multilayer perceptrons and embeds the reflection process into the network loss function. While NeRA demonstrates the feasibility of end-to-end radio map reconstruction, it primarily targets outdoor or open-space environments and does not explicitly incorporate semantic or material priors required for realistic indoor propagation modeling.

Building upon these insights, we extend our previous work EM DeepRay \cite{9771088}, which established a physics-informed and data-driven framework for indoor propagation learning on planar layouts. The proposed SenseRay-3D framework advances this line of research by extending the modeling to 3D prediction and introducing an end-to-end sensing-driven paradigm. It integrates RGB-D–based semantic reconstruction, physics-informed voxel representations encoding both occupancy and material-aware electromagnetic properties, and a SwinUNETR-based neural predictor. This design enables direct path-loss heatmap prediction from sensed 3D indoor scenes, bridging the gap between perception and propagation modeling, and achieving scalable, material-aware, and geometry-consistent indoor propagation modeling.

\section{System Model and Problem Formulation}

This section introduces the system model that governs signal propagation in indoor environments and formulates the corresponding prediction problem. Specifically, we describe the received-power relationship based on the Friis transmission equation \cite{Friis1946transmission}, decompose the total path-loss into free-space and environment-induced components, and define the learning task as estimating the spatial distribution of path-loss across the environment.

\subsection{System Model}
Consider a transmitter (Tx) located at $t=(x_t,y_t,z_t)$ in an indoor environment $\mathcal{S}$. The received power at a receiver (Rx) at location $v=(x_v,y_v,z_v) \in \mathcal{S}$ is given by the Friis transmission equation \cite{Friis1946transmission}. In linear scale, it is expressed as
\begin{equation}
    P_r(d) = P_t G_t G_r \left(\frac{\lambda}{4\pi d(t,v)}\right)^2,
    \label{eq:friis_linear}
\end{equation}
where $d(t,v)$ is the Euclidean distance between $t$ and $v$, $P_t$ is the transmit power, $G_t$ and $G_r$ are the Tx and Rx antenna gains, and $\lambda=c/f$ is the wavelength at carrier frequency $f$, $c$ is the speed of light.

Taking $10\log_{10}(\cdot)$ on both sides of \eqref{eq:friis_linear} gives
\begin{equation}
P_r(\mathrm{dB})
= 10\log_{10}\!\left(P_t G_t G_r 
\left(\frac{\lambda}{4\pi d(t,v)}\right)^2\right),
\end{equation}
which expands to
\begin{equation}
P_r(\mathrm{dB})
= P_t + G_t + G_r
+ 20\log_{10}\!\Big(\frac{\lambda}{4\pi d(t,v)}\Big).
\end{equation}
We then define the path-loss term as
\begin{align}
L(t,v)
&= 20\log_{10}\!\left(\frac{4\pi d(t,v)}{\lambda}\right) \nonumber\\[2pt]
&= -\,20\log_{10}\!\left(\frac{\lambda}{4\pi d(t,v)}\right),
\label{eq:path_loss}
\end{align}
so that the received power in decibel (dB) scale becomes
\begin{equation}
    P(t,v) = P_t + G_t + G_r - L(t,v),
    \label{eq:friis_db}
\end{equation}
which is the Friis transmission equation expressed in the decibel domain.

The total path-loss can be further decomposed as
\begin{equation}
    L(t,v) = L_{\text{fspl}}(t,v) + L_{\text{env}}(t,v),
    \label{eq:loss_decomp}
\end{equation}
where $L_{\text{fspl}}(t,v)$ denotes the FSPL, and $L_{\text{env}}(t,v)$ represents the environmental attenuation caused by reflection, diffraction, and transmission through obstacles.

Substituting $\lambda = c/f$ into~\eqref{eq:path_loss} gives
\begin{equation}
\begin{aligned}
L_{\text{fspl}}(t,v)
&= 20\log_{10}\!\left(\frac{4\pi f d(t,v)}{c}\right) \\[2pt]
&= 20\log_{10}\!\big(d(t,v)\big)
   + 20\log_{10}\!\big(f\big) \\
& + 20\log_{10}\!\!\left(\frac{4\pi}{c}\right),
\end{aligned}
\label{eq:fspl_deriv}
\end{equation}
where $d(t,v)$ is expressed in meters, $f$ in hertz. The constant term simplifies to
\[
20\log_{10}\!\left(\frac{4\pi}{c}\right) \approx -147.55~\text{dB},
\]
leading to the standard FSPL expression:
\begin{equation}
    L_{\text{fspl}}(t,v)
    = 20\log_{10}\!\big(d(t,v)\big)
    + 20\log_{10}\!\big(f\big)
    - 147.55.
    \label{eq:fspl}
\end{equation}
Unless otherwise stated, we make the following assumptions throughout this work:  
i) the transmit power $P_t$ is fixed and identical across all experiments,  
ii) both Tx and Rx employ omnidirectional antennas so that $G_t=G_r=0$ dB. 

\subsection{Problem Formulation}

From \eqref{eq:friis_db}–\eqref{eq:loss_decomp}, the total path-loss at location $v$ can be decomposed into two components: an analytically computable FSPL term $L_{\text{fspl}}(t,v)$ and an environmental path-loss $L_{\text{env}}(t,v)$. While $L_{\text{fspl}}(t,v)$ is directly determined by the transmitter–receiver distance and carrier frequency, $L_{\text{env}}(t,v)$ arises from complex propagation mechanisms such as reflection, diffraction, and transmission, and therefore cannot be expressed in closed form for realistic indoor environments.

Accordingly, the objective of indoor propagation modeling is to estimate the overall path-loss distribution $L(t,v)$ for all receiver locations $v \in \mathcal{S}$, given the indoor environment $\mathcal{S}$ and the Tx position $t$. In practice, this is achieved by learning to infer the environmental path-loss component $L_{\text{env}}(t,v)$ from $\mathcal{S}$ and $t$, and then reconstructing the total path-loss via \eqref{eq:loss_decomp}.

Formally, the problem can be represented as learning a mapping
\begin{equation}
    \mathcal{F}: (\mathcal{S}, t) 
    \mapsto \{L(t,v)\}_{v \in \mathcal{S}},
    \label{eq:mapping}
\end{equation}
where $\mathcal{F}$ represents the scene-dependent propagation function that characterizes the spatial distribution of total path-loss within the environment.

\section{Proposed Framework}

\begin{figure}[h]
    \centering
    \includegraphics[width=0.7\linewidth]{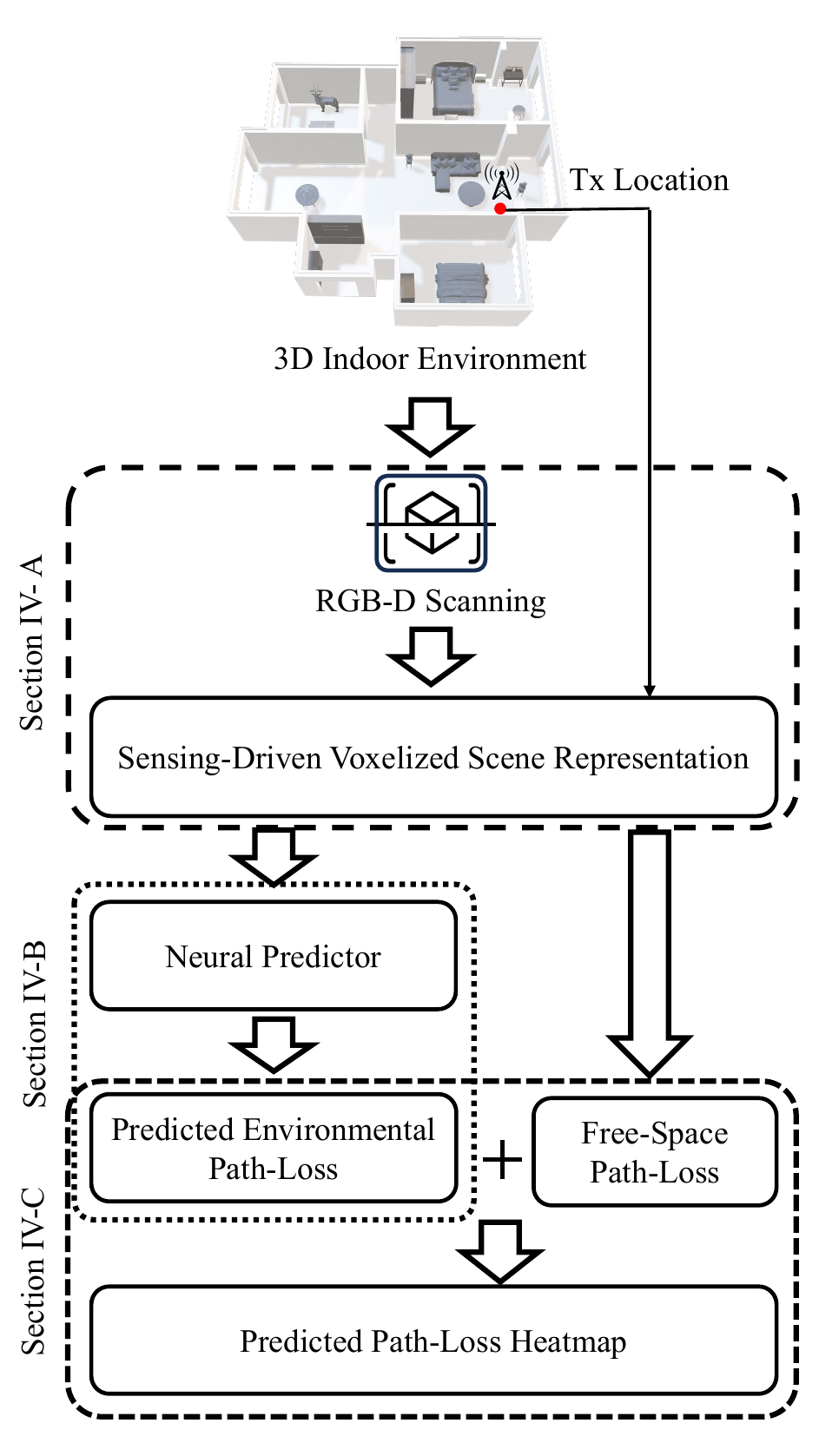}
    \vspace{-10pt}
    \caption{Schematic representation of SenseRay-3D}
    \label{fig:overview}
\end{figure}

In this section, we present SenseRay-3D, an end-to-end framework for indoor propagation modeling. The complete workflow is illustrated in Fig.~\ref{fig:overview}. The framework comprises three sequential stages: 
A. sensing-driven voxelized scene representation, 
B. SwinUNETR-based neural predictor, and 
C. path-loss reconstruction. 

The following subsections provide a detailed description of them.

\subsection{Sensing-Driven Voxelized Scene Representation}

The SenseRay-3D framework is initialized with the given indoor environment and Tx location. It proceeds to perform 3D sensing of the given indoor environment with RGB-D scanning, which jointly captures photometric appearance and geometric depth. These complementary modalities provide dense and metrically consistent descriptions of the scene, forming the foundation for subsequent semantic and physical modeling.

The RGB modality is first processed using semantic segmentation, assigning each pixel to a category label (e.g., wall, floor, ceiling, bed) that identifies the structural component it belongs to. The accompanying depth map provides metric distance information for every pixel, enabling accurate 3D reconstruction of the segmented environment.

Each RGB-D frame is back-projected into a semantically annotated point cloud in the camera coordinate system. According to the pinhole camera model \cite{Hartley_Zisserman_2004, szeliski2022computer, zhou2018open3d}, the 3D point $p_c$ corresponding to a pixel at image coordinate $(u,v)$ with depth value $z(u,v)$ and camera intrinsics $(f_x,f_y,c_x,c_y)$ is computed as
\begin{equation}
    p_c = 
    \left(
    \frac{(u-c_x)z(u,v)}{f_x},\,
    \frac{(v-c_y)z(u,v)}{f_y},\,
    z(u,v)
    \right),
    \label{eq:backprojection}
\end{equation}
which maps each pixel in the RGB-D image to a 3D point in the local camera frame.

To express all points in a unified global coordinate system, the local point cloud is then transformed by the camera-to-world extrinsic matrix $T_{cw}\in\mathbb{R}^{4\times4}$ as
\begin{equation}
    p_w = 
    \begin{bmatrix}
        x_w \\[2pt] y_w \\[2pt] z_w \\[2pt] 1
    \end{bmatrix}
    =
    T_{cw}
    \begin{bmatrix}
        p_c \\[2pt] 1
    \end{bmatrix},
    \label{eq:cam2world}
\end{equation}
where $p_w=(x_w,y_w,z_w)^\top$ denotes the 3D coordinates of the point 
in the world coordinate system. 
Here, 
\[
T_{cw} = 
\begin{bmatrix}
R_{cw} & t_{cw} \\[2pt]
0 & 1
\end{bmatrix}
\]
represents the camera-to-world transformation composed of a rotation matrix $R_{cw}$ 
and a translation vector $t_{cw}$, which together encode the pose of the camera 
relative to the global scene reference frame.

By aggregating the back-projected points from all camera views, we obtain a dense point cloud $\mathcal{P}=\{p_i\}_{i=1}^N$ covering the entire indoor environment. Each point is associated with its semantic label inherited from the segmentation maps, ensuring consistency between geometry and material attributes.

Each semantic category is then linked to its electromagnetic material properties through a predefined lookup table. For each category, parameters such as relative permittivity $\varepsilon_r$ and conductivity $\sigma$ are specified, following standardized material references. This semantic-to-material association enables the transformation of perceptual geometry into a physics-aware representation.

The fused point cloud is discretized into a voxel grid $\mathbf{V}\in\mathbb{R}^{D\times H\times W}$, where $D$, $H$, and $W$ denote the grid dimensions along the depth, height, and width, respectively. Each voxel $v\in\mathbf{V}$ is represented by a multi-channel feature vector incorporating geometry, material, and distance information relevant to radio propagation:

\subsubsection{Occupancy}
Occupancy encodes whether the voxel is free space or occupied by an object. It is defined as $o(v)=1$ if occupied and $o(v)=0$ otherwise.

\subsubsection{Reflection and Transmission Coefficients}
Electromagnetic waves interacting with indoor structures may undergo reflection, transmission, diffraction, or scattering \cite{9298918,1232163}. The relative strength of these mechanisms is governed by the ratio between the wavelength $\lambda$ and the characteristic dimensions of the objects in the environment. When walls, floors, or furniture are much larger than $\lambda$, incident waves are predominantly reflected or partially transmitted through the material. By contrast, diffraction mainly arises around small apertures or sharp edges when $\lambda$ is comparable to or larger than the obstacle size, while diffuse scattering requires surface irregularities on the order of the wavelength.  

In the frequency ranges of interest for 5G and beyond (sub-6~GHz and mmWave bands), most building structures are electrically large with respect to $\lambda$. Consequently, reflection at material boundaries and transmission through partitions become the most impactful mechanisms shaping the indoor propagation. Furthermore, their magnitudes are highly dependent on the material properties, which vary with frequency as specified in ITU-R recommendations \cite{ITU_R_P2040_3_2023}. 
To capture these effects in a learning framework, we explicitly encode reflection- and transmission-related features at the voxel level, thereby enabling the network to model dominant propagation phenomena in a material-aware manner.  

The electromagnetic response of each occupied voxel is then determined by its construction material. In our work, we assume all materials are non-ionized and non-magnetic ($\mu_r=1$). According to the ITU-R P.2040 recommendation \cite{ITU_R_P2040_3_2023}, the frequency-dependent material properties can be parameterized by four empirical constants $(a,b,c,d)$, which are tabulated for common indoor materials such as concrete, plasterboard, and wood. Based on these parameters, the relative permittivity $\varepsilon_r'$ and conductivity $\sigma$ at frequency $f$ (in GHz) are modeled as
\begin{equation}
    \varepsilon_r'(f) = a f_{\text{GHz}}^{b}, \quad
    \sigma(f) = c f_{\text{GHz}}^{d}.
\end{equation}

Given $(\varepsilon_r',\sigma)$, the complex permittivity of the 
material is expressed as
\begin{equation}
    \varepsilon = \varepsilon_r \varepsilon_0 
    = \left(\varepsilon_r' - j\frac{\sigma}{\omega \varepsilon_0}\right)\varepsilon_0,
\end{equation}
where $\varepsilon_0$ is the vacuum permittivity and $\omega=2\pi f$ is the angular frequency. Equation (13) combines both dielectric storage (real part) and conductive loss (imaginary part) into a unified representation of the material’s electromagnetic response. This complex permittivity directly determines how an electromagnetic wave interacts with the medium, as it governs the intrinsic impedance of the material. 

Accordingly, the wave impedance of a homogeneous isotropic medium with relative permittivity $\varepsilon_r$, conductivity $\sigma$, and free space permeability $\mu_0$, is given by
\begin{equation}
    \eta = \sqrt{\frac{j\omega\mu_0}{\sigma + j\omega \varepsilon_r'\varepsilon_0}},
\end{equation}
which is derived from Maxwell's equations for plane wave propagation in lossy media \cite{balanis2012advanced,cheng2014field}. Based on this impedance and the Fresnel equations, the reflection coefficient $\Gamma$ and transmission coefficient $T$  at normal incidence can be estimated from the boundary conditions of Maxwell’s equations \cite{balanis2012advanced} as
\begin{equation}
    \Gamma = \frac{\eta - \eta_0}{\eta + \eta_0}, \quad
    T = \frac{2\eta}{\eta+\eta_0},
\end{equation}
with $\eta_0$ denoting the intrinsic impedance of free space. Finally, their magnitudes are converted into dB scale and assigned as voxel-wise features:
\begin{equation}
    \rho(v) = 20\log_{10}|\Gamma|, \quad
    \tau(v) = 20\log_{10}|T|,
\end{equation}
where $\rho(v)$ and $\tau(v)$ denote the reflection-related and transmission-related feature channels, respectively.
In this way, the ITU material parameters $(a,b,c,d)$ are systematically mapped to voxel-level reflection and transmission features, providing the neural network with explicit material-dependent propagation cues. 

\subsubsection{Distance}
For each voxel center $v=(x_v,y_v,z_v)$, the Euclidean distance to the given Tx $t=(x_t,y_t,z_t)$ is computed as
\begin{equation}
    d(t,v) = \sqrt{(x_v-x_t)^2+(y_v-y_t)^2+(z_v-z_t)^2}.
    \label{eq:distance}
\end{equation}
This channel encodes the fundamental large-scale attenuation of wireless signals, as the FSPL is directly determined by distance and frequency. By explicitly including $d(t,v)$ as an input, the model is provided with a physics-based prior that captures the dominant distance-dependent trend.

\begin{figure*}[t]
    \centering
    \includegraphics[width=0.8\linewidth]{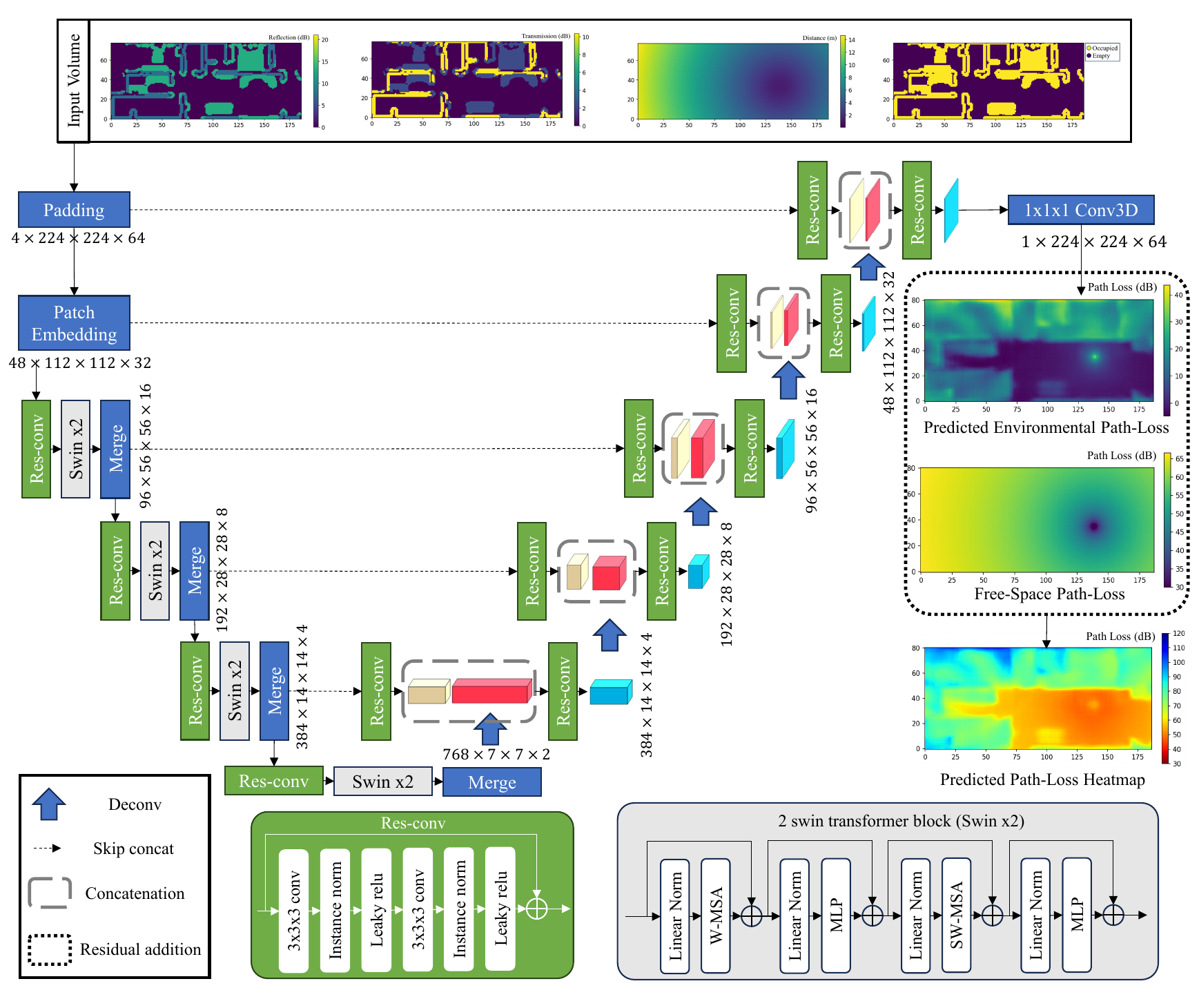}
    \caption{Detailed architecture of the SwinUNETR-based predictor used in SenseRay-3D, showing voxelized input features, hierarchical encoding and decoding, and reconstruction of the path-loss heatmap.}
    \label{fig:network}
\end{figure*}

\subsection{SwinUNETR-Based Neural Predictor}

We adopt the SwinUNETR architecture \cite{he2023swinunetrv2} as the backbone of SenseRay-3D. It is a 3D U-Net variant in which conventional convolutional encoders are replaced by hierarchical Swin Transformer blocks, combining U-Net’s local feature aggregation with Transformer-based long-range dependency modeling \cite{swintransmformer,swinunetr,Tang_2022_CVPR,UNETR}. This hybrid design is particularly beneficial for radio propagation modeling, where the received power depends not only on its immediate surroundings but also on reflections and transmissions from distant regions of the environment.

The choice of SwinUNETR is motivated by two considerations. First, indoor propagation involves multi-scale interactions. Large-scale structures such as walls and partitions determine the dominant propagation paths, whereas fine-scale details at material boundaries introduce additional reflections and diffractions. A model capable of capturing both global and local dependencies is therefore required. Second, the voxelized scene representation forms a dense 3D grid that naturally aligns with the volumetric input format of SwinUNETR. Originally designed for volumetric medical image segmentation, SwinUNETR has demonstrated strong performance in dense 3D prediction tasks, which we extend here to the domain of radio propagation modeling.

The overall architecture of the proposed SwinUNETR-based neural predictor is illustrated in Fig.~\ref{fig:network}. The model operates on the voxelized input tensor 
\[
    \mathbf{X}\in \mathbb{R}^{4\times D\times H\times W},
\]
where the four channels correspond to (i) occupancy, (ii) reflection coefficient, (iii) transmission coefficient, and (iv) transmitter–voxel distance. The input is first center-padded to a fixed resolution of $4\times224\times224\times64$ to accommodate variable scene dimensions. A patch embedding layer divides the input volume into non-overlapping 3D patches, which are linearly projected into an embedding space of dimension 48.

The encoder consists of four hierarchical stages, each composed of a residual convolutional block followed by two Swin Transformer blocks (denoted as Swin×2) and a patch merging operation that halves the spatial resolution while doubling the channel dimension. This design progressively expands the receptive field and captures long-range spatial dependencies across the scene. The bottleneck operates at the coarsest resolution ($768\times7\times7\times2$) and integrates global contextual information that spans the entire environment.

The decoder mirrors the encoder through a series of 3D deconvolution layers that progressively upsample the feature maps. At each resolution level, skip connections concatenate the upsampled features with the corresponding encoder outputs, followed by residual convolutional refinement. This structure preserves fine-grained geometric information while maintaining the global propagation context. A final $1\times1\times1$ convolutional projection maps the decoder output to a single-channel residual volume
\[
    \hat{\mathbf{R}}\in \mathbb{R}^{1\times D\times H\times W},
\]
representing the predicted environment-induced attenuation relative to free-space propagation.

\begin{figure*}
    \centering
    \includegraphics[width=0.75\linewidth]{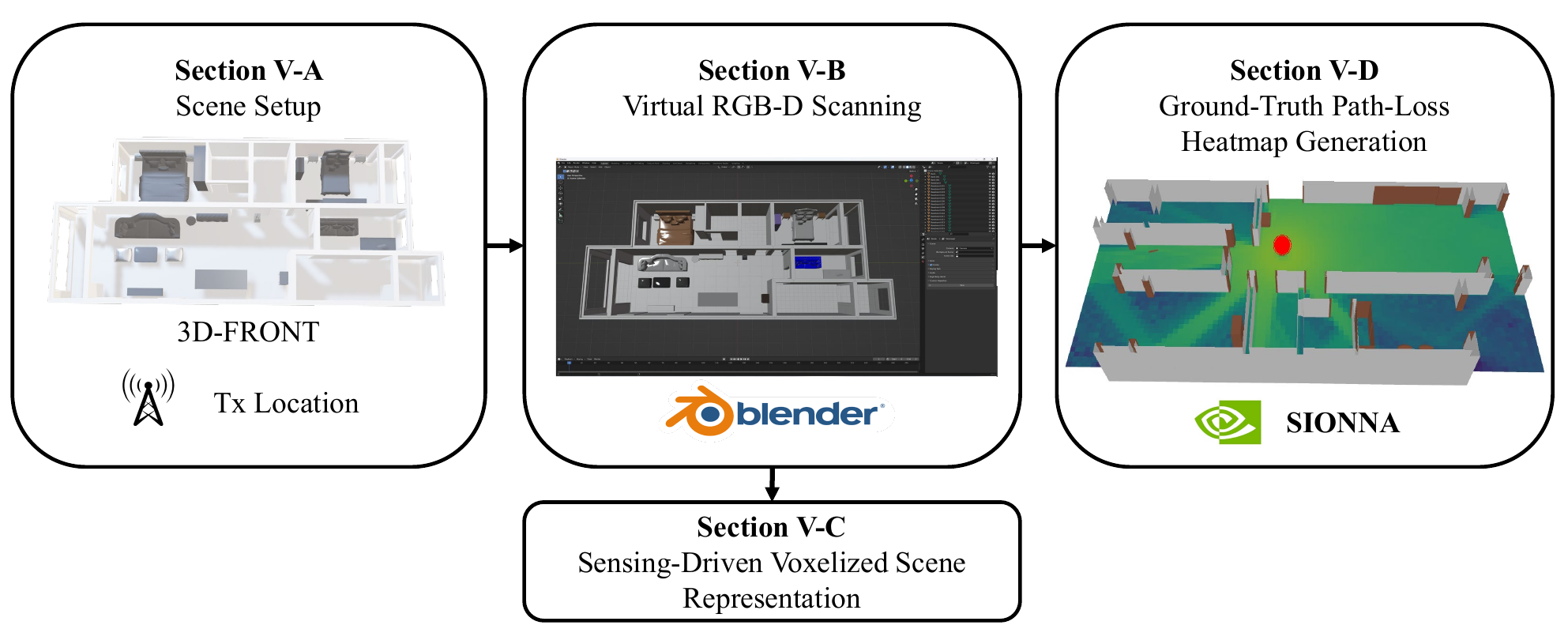}
    \caption{Overview of the proposed dataset generation process}
    \label{fig:dataset_overview}
\end{figure*}

\subsection{Path-Loss Reconstruction}

The proposed network is trained in a supervised manner to predict the total path-loss $L_(t,v)$ from voxelized scene features. To achieve this, a residual learning strategy is adopted, where the model estimates the environmental path-loss component $L_\text{env}(t,v)$ that complements the analytically derived FSPL term $L_{\text{fspl}}(t,v)$. This design enhances numerical stability and ensures physical consistency with the distance-dependent propagation characteristics.

Let $\hat{\mathbf{R}}=f_{\theta}(\mathbf{X})$ denote the network prediction parameterized by $\theta$. The final path-loss heatmap is reconstructed as
\begin{equation}
    \hat{L}(t,v) = L_{\text{fspl}}(t,v) + \hat{L}_{\text{env}}(t,v),
    \label{eq:reconstruction}
\end{equation}
where $\hat{L}_{\text{env}}(t,v)=\hat{\mathbf{R}}(v)$ represents the predicted residual attenuation. The ground-truth labels $L_{\text{env}}(t,v)$ are obtained from deterministic electromagnetic simulations, and the model is optimized to minimize the voxel-wise prediction error.

This residual formulation offers two major advantages. First, by isolating the analytically determined FSPL, the network is relieved from modeling trivial distance-dependent variations and can focus on complex interactions such as reflection and transmission. Second, the residual constraint naturally preserves physical realism, ensuring that predicted path-loss remains consistent with fundamental propagation principles while achieving high spatial resolution and accuracy.

\section{Dataset Generation}
In this section, we construct a comprehensive dataset to validate the effectiveness of the proposed method and to provide a benchmark for future research on indoor propagation modeling. The dataset consists of paired voxelized scene features and ground-truth path-loss heatmaps for supervised learning. 
 
As illustrated in Fig.~\ref{fig:dataset_overview}, the overall data generation pipeline consists of four main stages: 
A. scene setup, 
B. virtual RGB-D scanning, 
C. sensing-driven voxelized scene representation, and 
D. ground-truth path-loss heatmap generation. 


\subsection{Scene Setup}
The dataset construction is based on the 3D-FRONT dataset \cite{fu20213d}, which provides a large collection of 3D indoor layouts with detailed furniture and structural elements. 
Each scene includes the original computer-aided design (CAD) models, offering precise geometric representations of indoor structures. This dataset is well-suited for propagation modeling as it contains diverse layouts and heterogeneous materials, enabling the construction of realistic indoor environments. In total, 56 furnished apartments are selected, spanning a range of floor areas from compact units of 70–80 m$^2$ to large apartments of approximately 250 m$^2$. Tx positions are randomly sampled within each apartment. Specifically, duplicate positions are avoided to prevent data redundancy. Specifically, about 20–50 Tx locations are sampled in each indoor environment, depending on the scene size.

\subsection{Virtual RGB-D Scanning}
\begin{figure*}[t]
    \centering
    \begin{subfigure}[b]{0.26\textwidth}
        \centering
        \includegraphics[width=\textwidth]{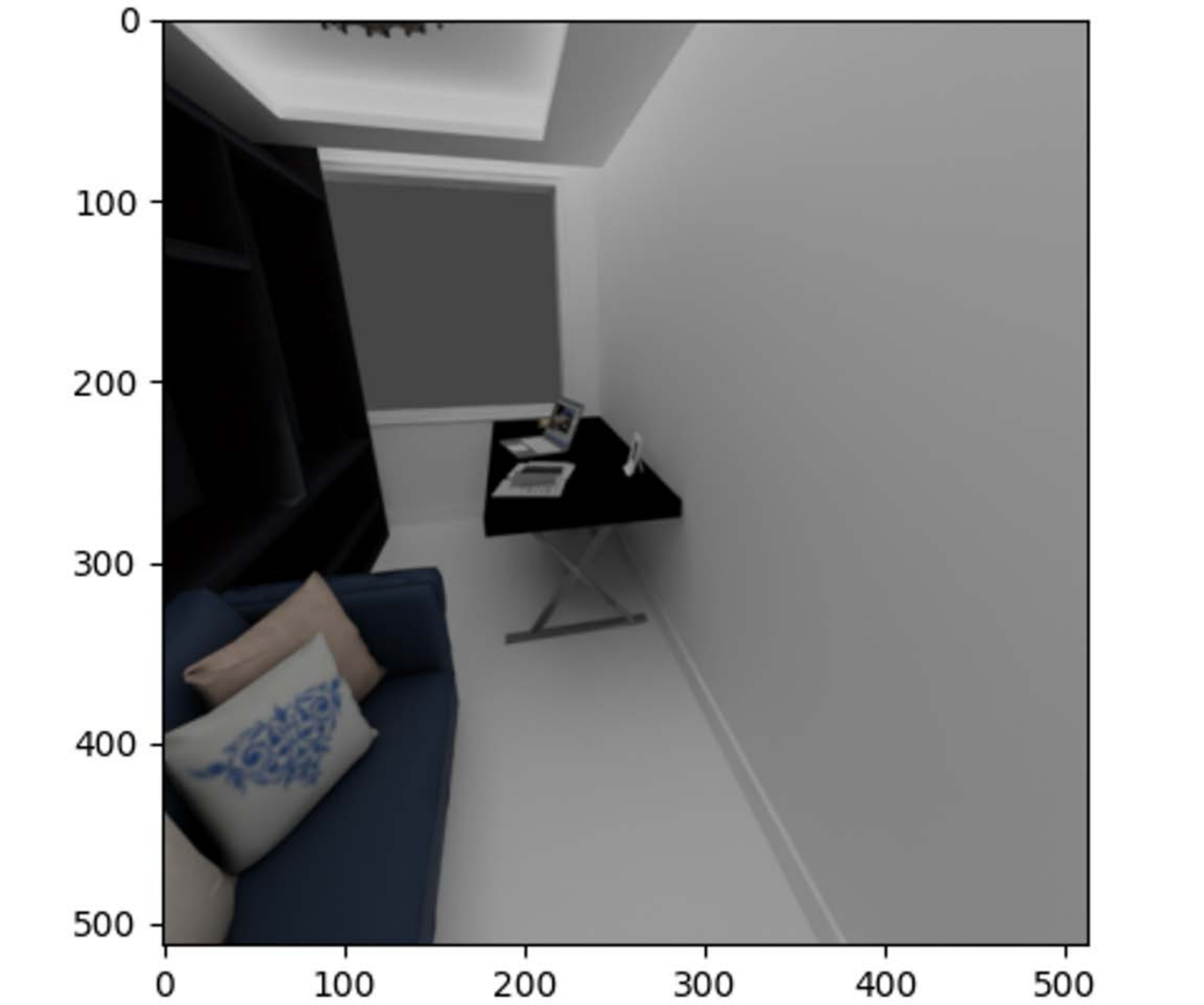}
        \vspace{-15pt}
        \caption{}
        \label{fig:RGB image}
    \end{subfigure}
    \centering
    \begin{subfigure}[b]{0.26\textwidth}
        \centering
        \includegraphics[width=\textwidth]{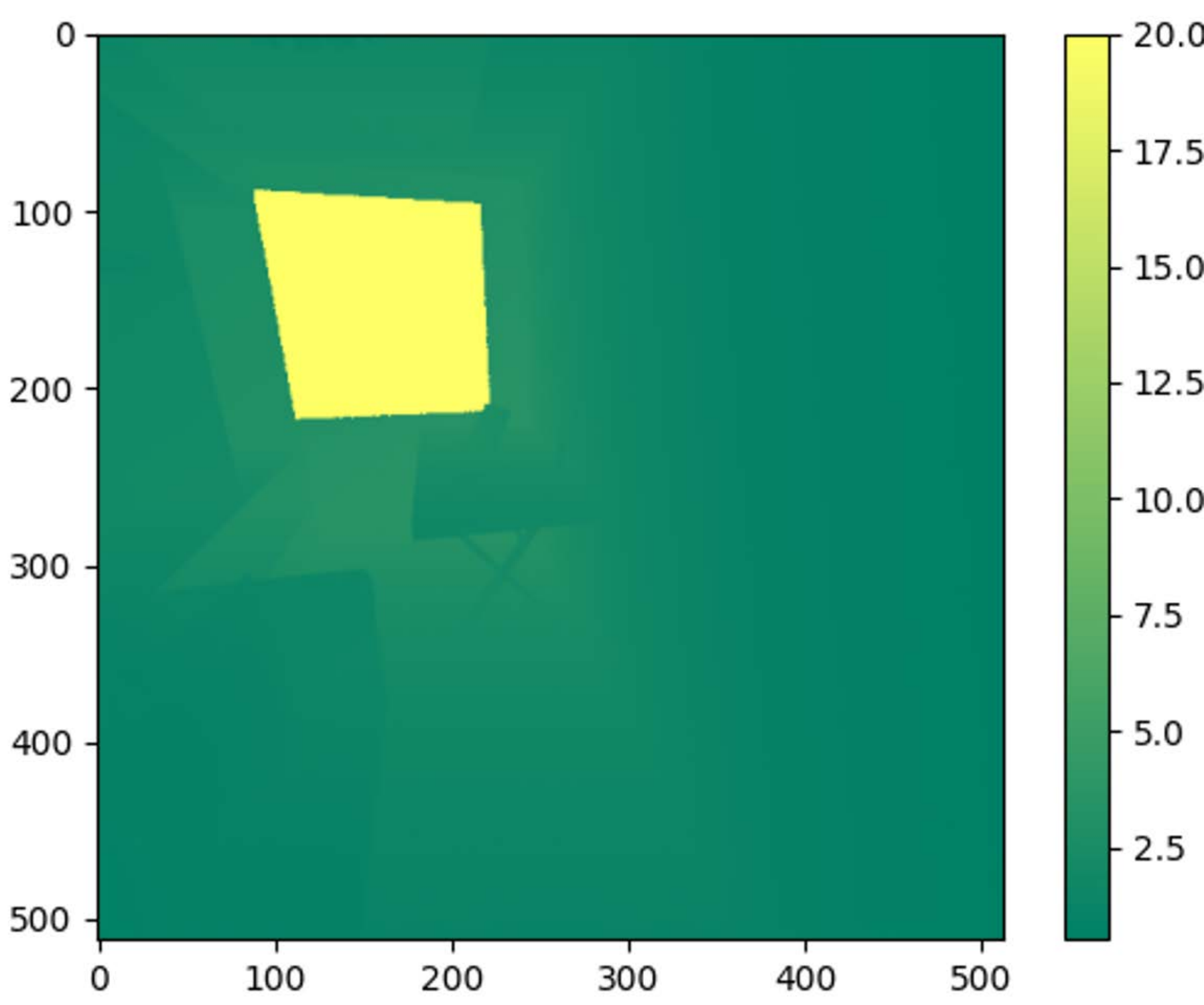}
        \vspace{-15pt}
        \caption{}
        \label{fig:Depth image}
    \end{subfigure}
    \centering
    \begin{subfigure}[b]{0.26\textwidth}
        \centering
        \includegraphics[width=\textwidth]{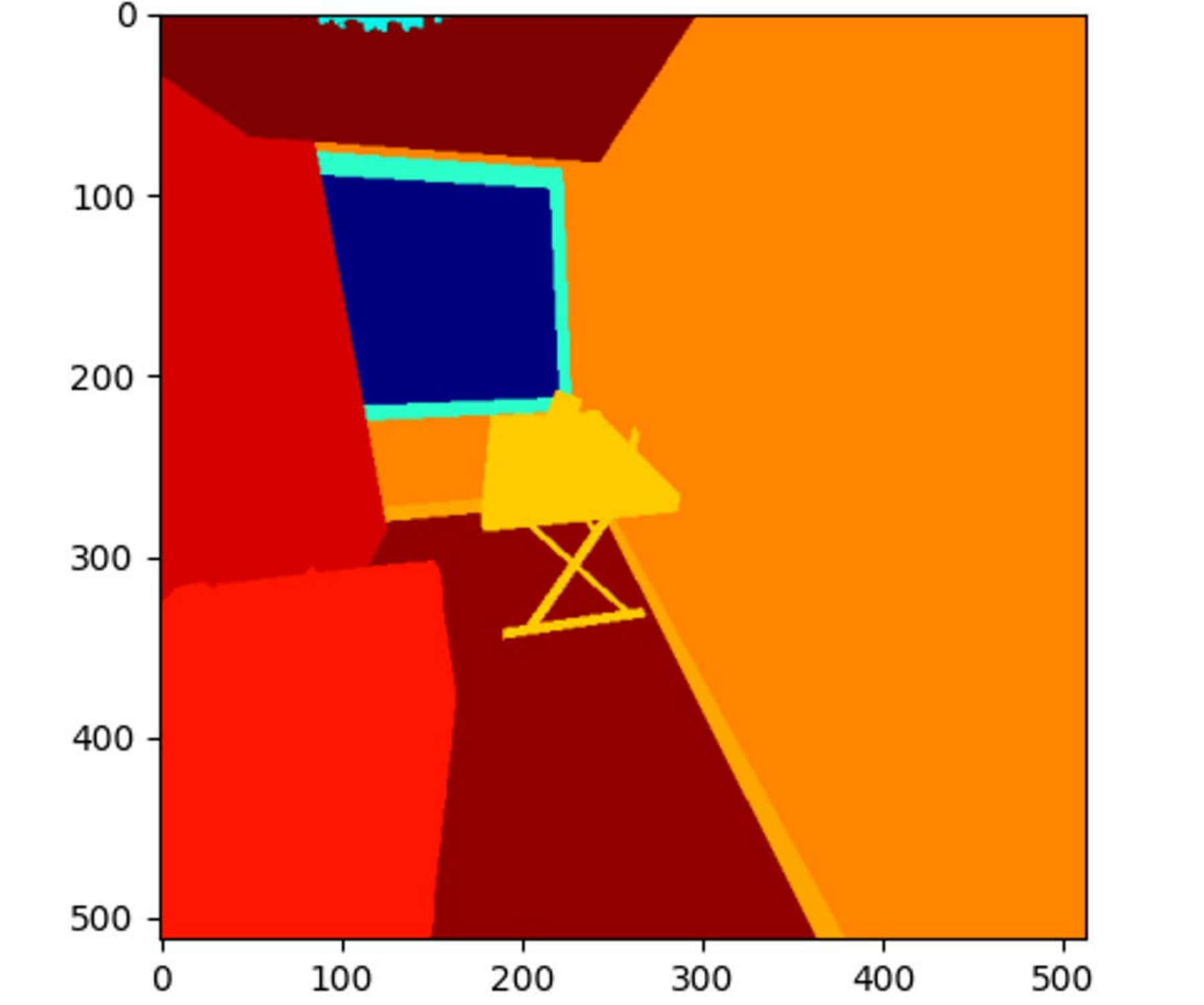}
        \vspace{-15pt}
        \caption{}
        \label{fig:Semantic segmentation map}
    \end{subfigure}
    \caption{Virtual RGB-D scanning results: (a) RGB image, (b) depth map, and (c) semantic segmentation map.}
    \label{fig:rendered}
\end{figure*}

\begin{table*}[!t]
\centering
\begin{tabular}{lcccccccc}
\toprule
Material (3.5GHz) & {$a$} & {$b$} & {$c$} & {$d$} & {$\varepsilon_r'$} & {$\sigma$ [S/m]} & {$\rho$ [dB]} & {$\tau$ [dB]} \\
\midrule
Vacuum       & 1.000 & 0.0 & 0.0000 & 0.0000 & 1.000 & 0.00000 & \(-\infty\) & 0.00 \\
Concrete     & 5.240 & 0.0 & 0.0462 & 0.7822 & 5.240 & 0.12309 & -7.49 & -10.36 \\
Brick        & 3.910 & 0.0 & 0.0238 & 0.1600 & 3.910 & 0.02908 & -6.66 & -3.77 \\
Plasterboard & 2.730 & 0.0 & 0.0085 & 0.9395 & 2.730 & 0.02758 & -14.12 & -3.07 \\
Wood         & 1.990 & 0.0 & 0.0047 & 1.0718 & 1.990 & 0.01799 & -13.30 & -2.41 \\
Ceiling\_board  & 1.480 & 0.0 & 0.0011 & 1.0750 & 1.480 & 0.00423 & -21.00 & -0.61 \\
Marble       & 7.074 & 0.0 & 0.0055 & 0.9262 & 7.074 & 0.01755 & -5.78 & -2.86 \\
Metal        & 1.000 & 0.0 & 107.00 & 0.0000 & 1.000 & 107.000 & 0.00 & \(-\infty\) \\
\bottomrule
\end{tabular}
\caption{Electromagnetic Parameters and Derived Reflection/Transmission Coefficients of Building Materials at 3.5GHz}
\label{tab:material}
\end{table*}

To emulate realistic sensing conditions, we simulate the scanning process using Blender and  BlenderProc \cite{Denninger2023}, which render RGB-D scans from multiple virtual camera viewpoints within each apartment. Each rendered frame contains:
\begin{itemize}
    \item An RGB image rendered under physically consistent lighting conditions.
    \item A depth map, aligned to the RGB frame, providing metric distances from the camera plane to visible surfaces.
    \item A semantic segmentation map with category labels. 
    \item Camera intrinsics and extrinsics, including focal length, principal point, and the full camera-to-world transformation matrix.
\end{itemize}
As illustrated in Fig.~\ref{fig:rendered}, the combination of RGB, depth, and semantic information reproduces the modalities of a real RGB-D sensor and serves as input to the voxelization process. In practical deployment, semantic segmentation maps are typically obtained by applying pretrained segmentation algorithms, such as Segment Anything \cite{kirillov2023segment}, on captured RGB images.

\subsection{Sensing-Driven Voxelized Scene Representation}

As described in Section IV-A, each rendered frame is back-projected into 3D space according to the intrinsic camera parameters using the transformation in \eqref{eq:backprojection}. The resulting local point clouds are then mapped into the global coordinate system through the camera-to-world transformation defined in \eqref{eq:cam2world}. A dense and semantically annotated global point cloud is obtained for each apartment by aggregating the back-projected points from all camera viewpoints. 

The fused point cloud is discretized into a voxel grid $\mathbf{V}\in\mathbb{R}^{D\times H\times W}$ with a voxel size of $0.1$~m. Each voxel is assigned four feature channels as defined in Section~IV-A: occupancy, reflection coefficient, transmission coefficient, and transmitter–voxel distance. The material-dependent parameters $a$, $b$, $c$, $d$, $\varepsilon_r'$, $\sigma$, $\rho$, and $\tau$ are derived from the ITU-R P.2040 recommendation~\cite{ITU_R_P2040_3_2023} and summarized in Table~\ref{tab:material}. The transmitter–voxel distance is computed using~(\ref{eq:distance}) based on the sampled Tx positions, while FSPL for each voxel can be obtained from~(\ref{eq:fspl}). These quantities together form the model input tensor $\mathbf{X}\in\mathbb{R}^{4\times D\times H\times W}$. 

\subsection{Ground-Truth Path-Loss Heatmap Generation}

The voxelized features described above require corresponding path-loss heatmaps as ground truth. To obtain this, deterministic modeling is employed for radio propagation modeling. In our previous work EM DeepRay \cite{9771088}, we utilized Ranplan Professional \cite{ranplan}, a commercial-grade ray-tracing simulator developed by Ranplan Wireless. Ranplan Professional provides a robust and industry-validated propagation engine capable of modeling both indoor and outdoor environments and has been widely adopted in practical network planning. To ensure reproducibility, open accessibility, and broader compatibility with emerging learning-based research pipelines, we adopt the Sionna RT \cite{sionnart} module from NVIDIA Sionna \cite{sionna} in this work. Sionna is an open-source, differentiable framework for end-to-end wireless communication system modeling, providing GPU-accelerated implementations of key physical-layer components. Its ray-tracing engine integrates with Mitsuba for physically accurate light and radio propagation and includes material models compliant with ITU-R and 3GPP specifications, thereby enabling efficient and reproducible electromagnetic scene modeling within the learning pipeline.

The 3D geometry of each scene is directly derived from the 3D-FRONT dataset \cite{fu20213d}, consistent with the virtual scanning process described in Section~V-B. The resulting scene is exported in a Mitsuba-compatible XML format, which serves as an input scene description for Sionna RT. Each object in the XML file is assigned electromagnetic material properties in accordance with ITU-R P.2040 \cite{ITU_R_P2040_3_2023}, enabling Sionna RT to perform realistic simulations of reflection, transmission, and scattering.

For each Tx, the solver computes the received power distribution over a uniform grid while accounting for multi-path effects. Tx is placed at sampled positions in the scene and modeled as omnidirectional point sources operating at $f=3.5$ GHz, corresponding to the 5G N78 band \cite{3GPP_TS_38_101_1_V18_8_0}. The outputs are path-loss heatmaps with a spatial resolution of 0.1 m x 0.1 m in the horizontal plane.   

To account for realistic user equipment heights and vertical propagation effects, path-loss heatmaps are extracted on horizontal planes ranging from 0.6 m to 1.6 m above the floor, representing the typical height range of handheld devices for seated and standing users, with 0.1 m spacing. Due to occasional full blockage or insufficient ray sampling, certain grid cells lack received-power data. These missing samples are reconstructed through bilinear interpolation followed by nearest-neighbor filling to maintain spatial consistency and completeness of the heatmaps. The result received power values are then converted into dB scale.

\begin{figure*}[t]
    \centering
    \begin{subfigure}[b]{0.4\textwidth}
        \centering
        \includegraphics[width=\textwidth]{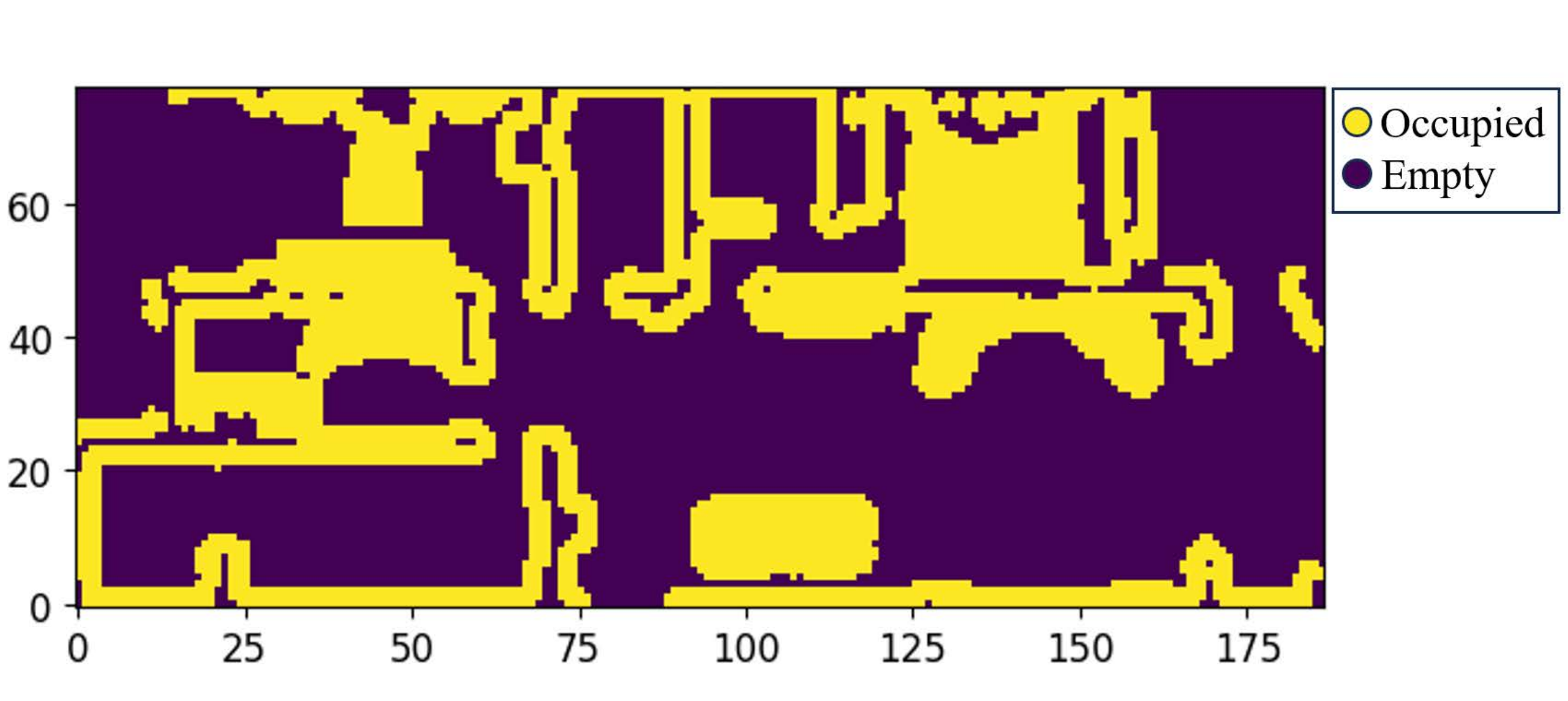}
        \caption{}
        \label{fig:occupancy}
    \end{subfigure}
    \centering
    \begin{subfigure}[b]{0.4\textwidth}
        \centering
        \includegraphics[width=\textwidth]{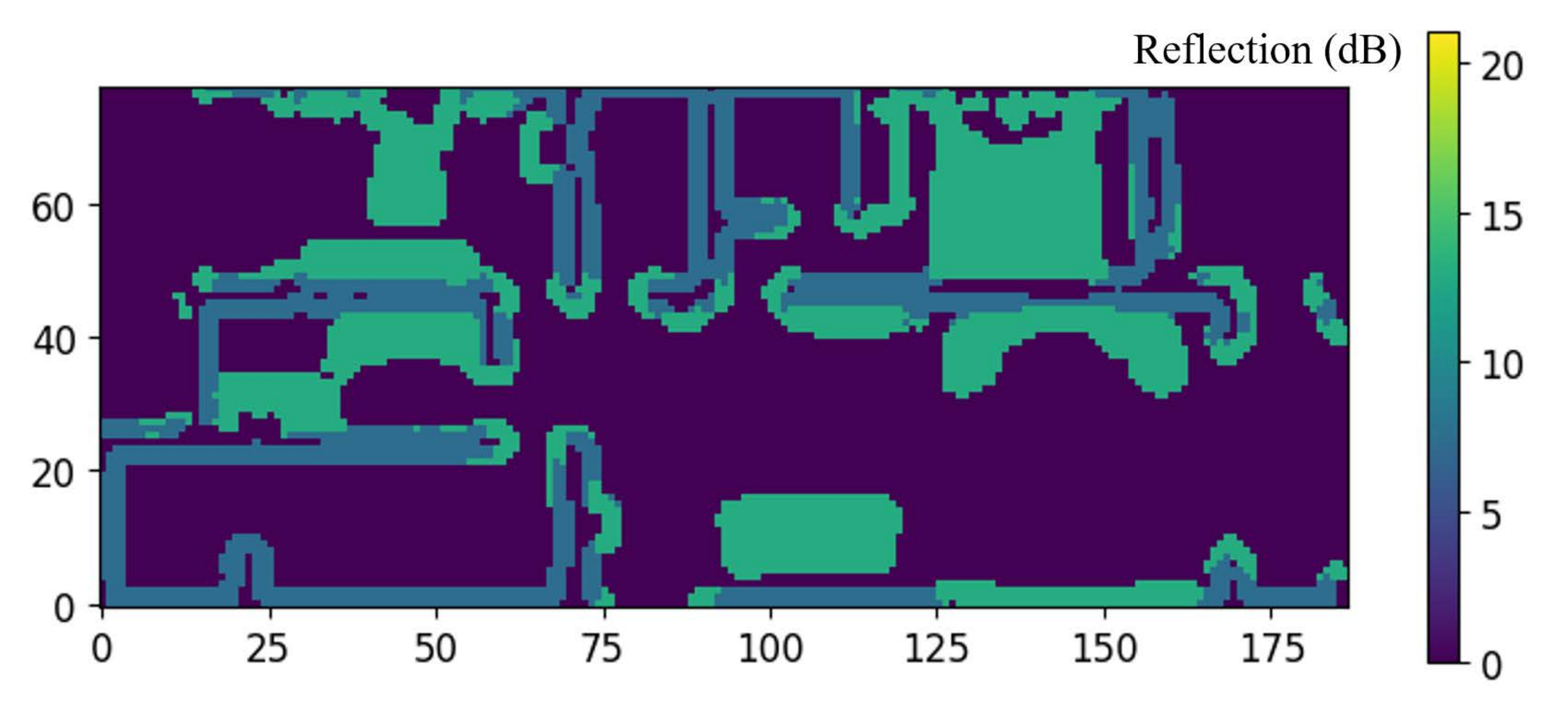}
        \caption{}
        \label{fig:reflection}
    \end{subfigure}
    \centering
    \begin{subfigure}[b]{0.4\textwidth}
        \centering
        \includegraphics[width=\textwidth]{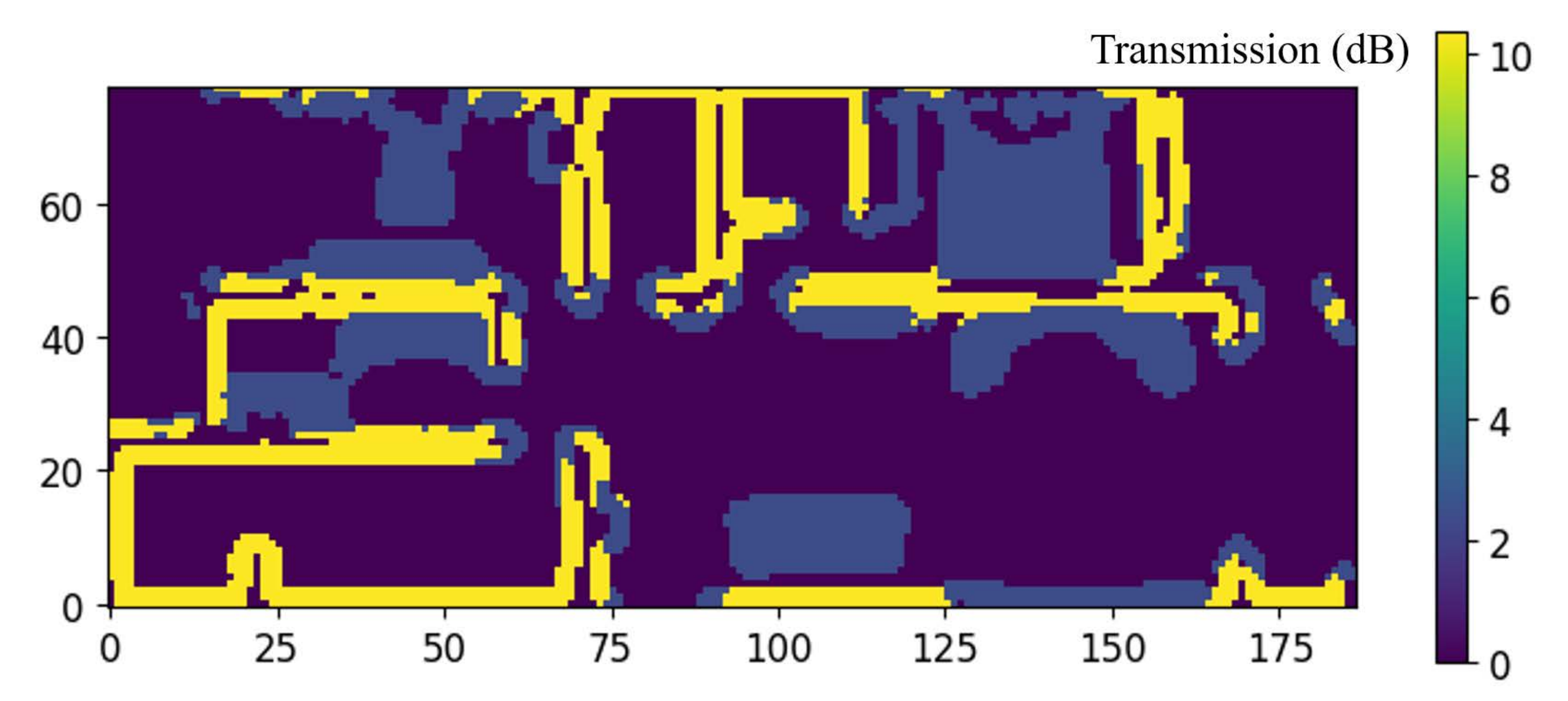}
        \caption{}
        \label{fig:transmission}
    \end{subfigure}
    \centering
    \begin{subfigure}[b]{0.4\textwidth}
        \centering
        \includegraphics[width=\textwidth]{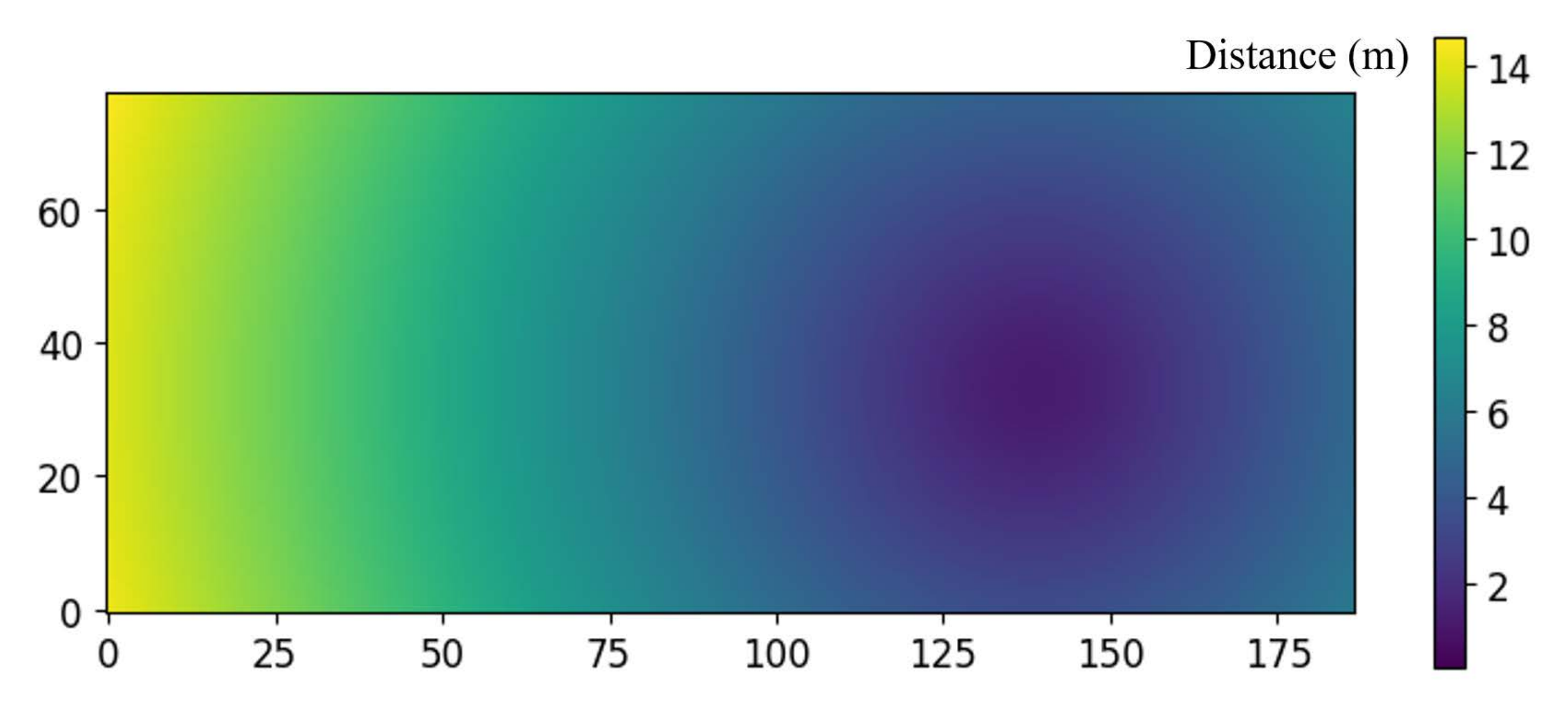}
        \caption{}
        \label{fig:distance}
    \end{subfigure}
    \centering
    \begin{subfigure}[b]{0.4\textwidth}
        \centering
        \includegraphics[width=\textwidth]{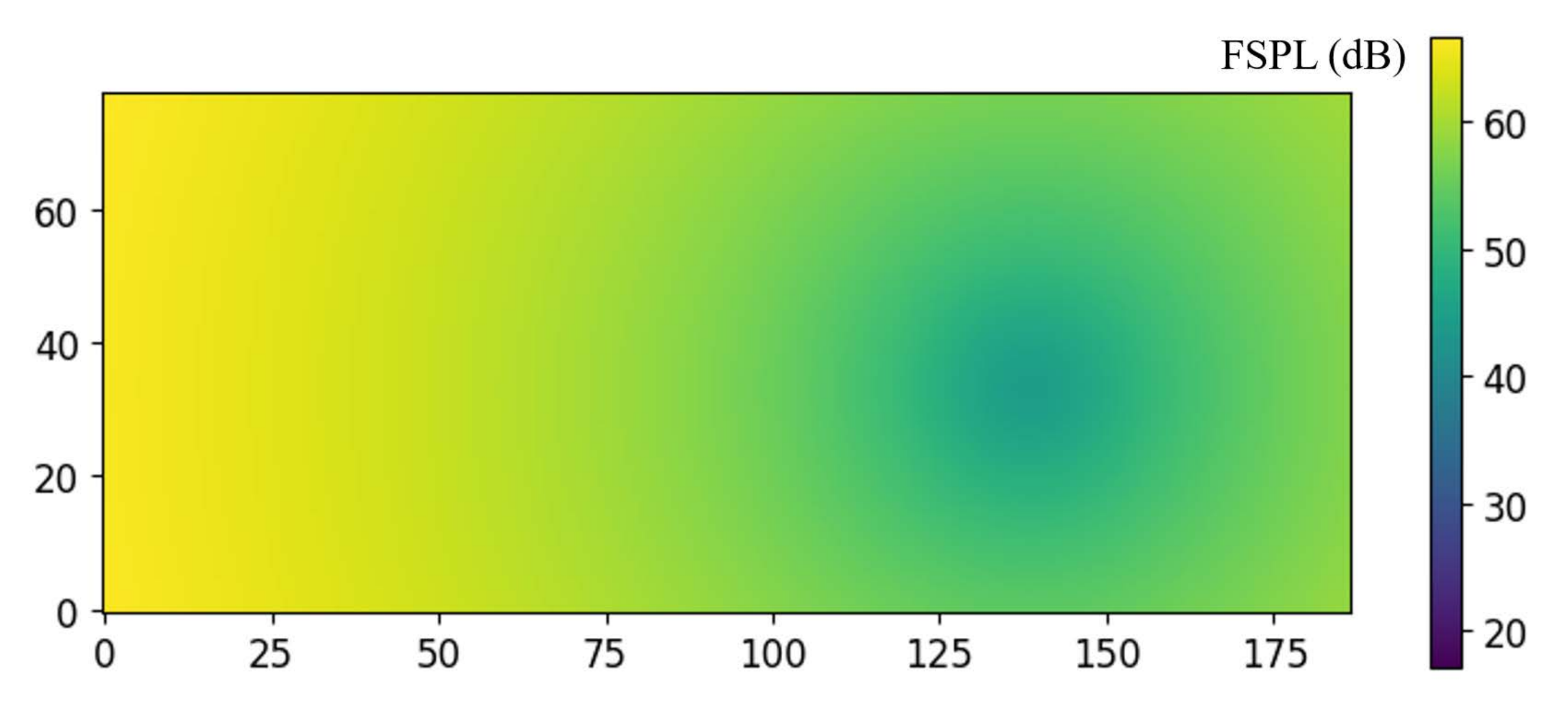}
        \caption{}
        \label{fig:fspl}
    \end{subfigure}
    \centering
    \begin{subfigure}[b]{0.4\textwidth}
        \centering
        \includegraphics[width=\textwidth]{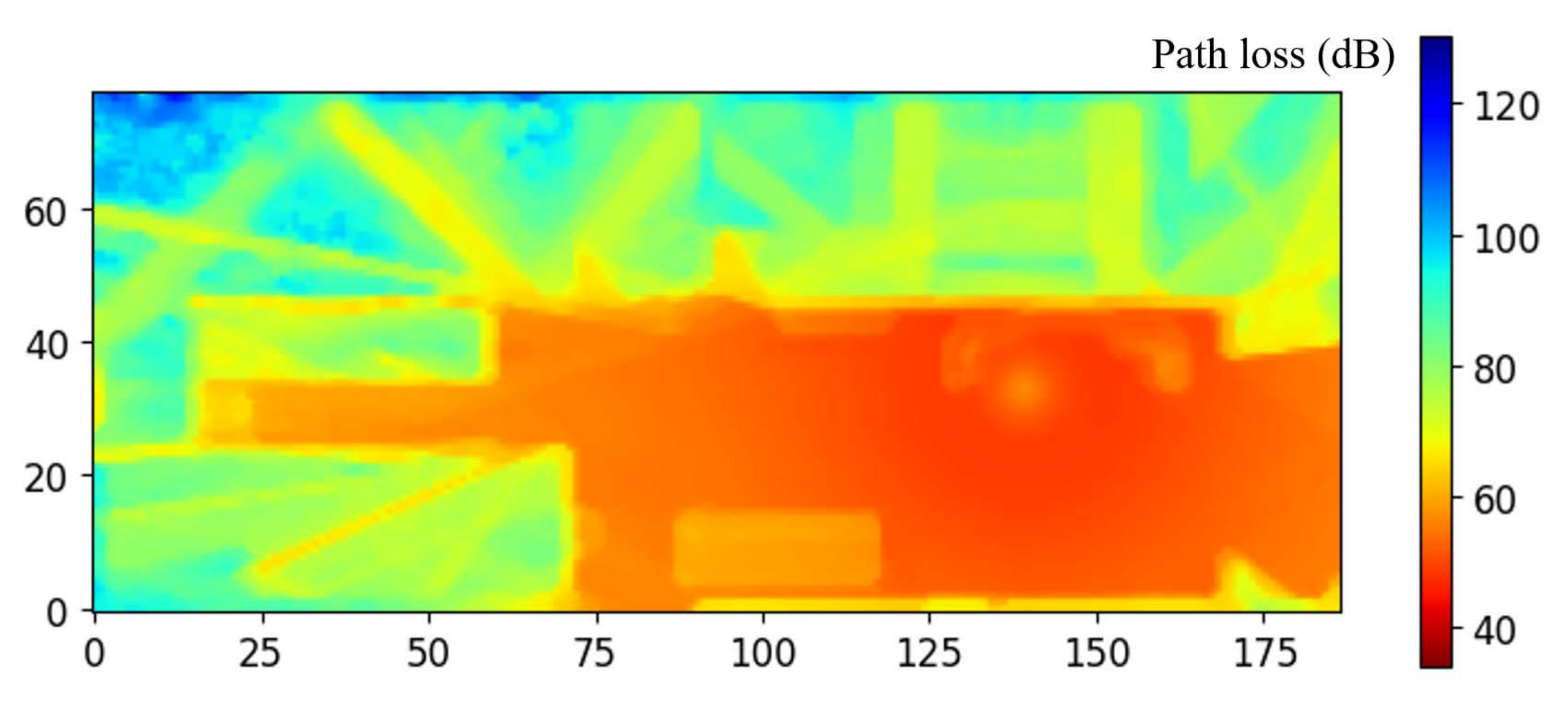}
        \caption{}
        \label{fig:radiomap}
    \end{subfigure}
    \caption{Illustration of the dataset representation used for 
supervised training. Each sample contains voxelized input features: 
(a) occupancy grid describing scene geometry, 
(b) reflection feature (dB) derived from ITU-based material parameters, 
(c) transmission feature (dB), and 
(d) transmitter–voxel distance field (m). In addition, 
(e) the FSPL (dB) serves as the baseline, while 
(f) the ground-truth path-loss heatmap (dB) obtained from Sionna.}
    \label{fig:dataset}
\end{figure*}

All generated samples are stored in the HDF5 format \cite{hdf5}, which is widely adopted in scientific computing for managing large-scale multidimensional data. The HDF5 format provides several advantages over flat binary or text-based storage \cite{hdf5}:  
\begin{enumerate}
    \item It supports complex, high-rank arrays and heterogeneous data types within a single file; 
    \item It enables efficient I/O through chunked storage, compression, and optional parallel access;  
    \item Its self-describing nature guarantees long-term portability and cross-platform usability;
    \item It integrates seamlessly with popular scientific tools such as MATLAB, Python, and Pytorch.     
\end{enumerate}
  
These properties make HDF5 particularly suitable for handling the voxelized tensors and path-loss heatmaps in our dataset. Each entry in the dataset corresponds to one transmitter configuration and includes:  
\begin{itemize}
    \item Occupancy grid (binary);
    \item Reflection coefficient (in dB);
    \item Transmission: coefficient (in dB);
    \item Transmitter-distance voxel (in m);
    \item Tx coordinate (in m);
    \item FSPL baseline (in dB);
    \item Ground truth path-loss heatmap (in dB).
\end{itemize}

All tensors are stored in float32 precision with optional gzip compression to balance storage efficiency and numerical accuracy. The unified format enables direct integration with PyTorch and TensorFlow data loaders for efficient batching and augmentation. An example dataset entry with all feature channels and the ground-truth is illustrated in Fig.~\ref{fig:dataset}.

\begin{figure}[t]
    \centering
    \begin{subfigure}[b]{0.4\textwidth}
        \centering
        \includegraphics[width=\textwidth]{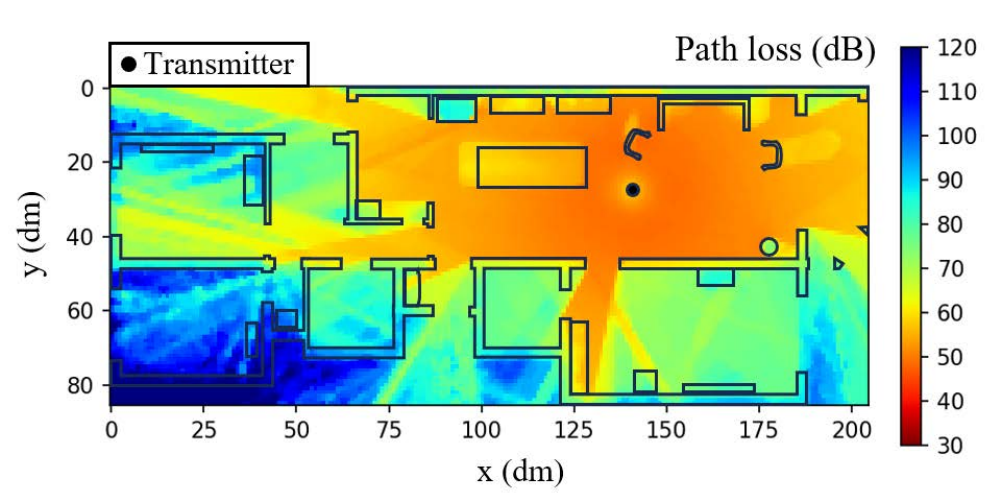}
        \vspace{-20pt}
        \caption{}
        \label{fig:test_gt_0_6}
    \end{subfigure}
    \vspace{-1.9pt}
    \vfill
    \begin{subfigure}[b]{0.4\textwidth}
        \centering
        \includegraphics[width=\textwidth]{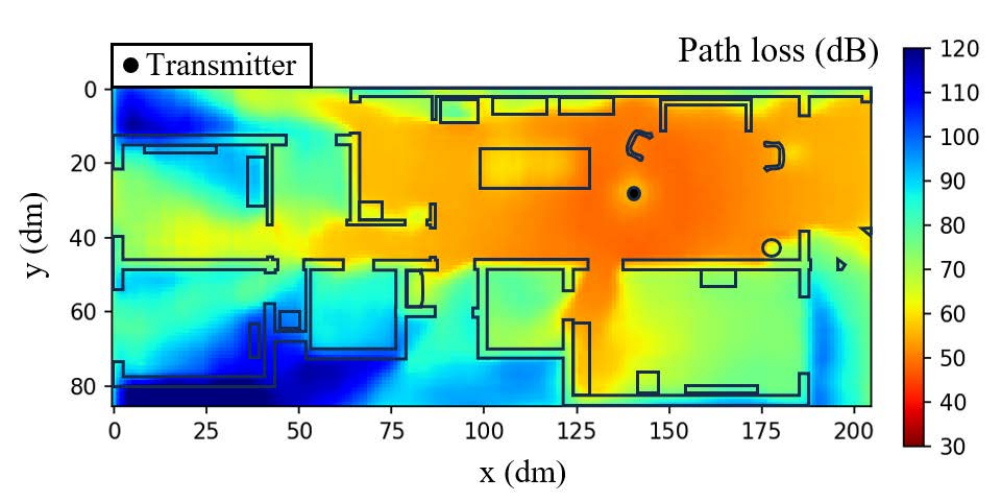}
        \vspace{-20pt}
        \caption{}
        \label{fig:test_pred_0_6}
    \end{subfigure}
    \vspace{-1.9pt}
    \vfill
    \begin{subfigure}[b]{0.4\textwidth}
        \centering
        \includegraphics[width=\textwidth]{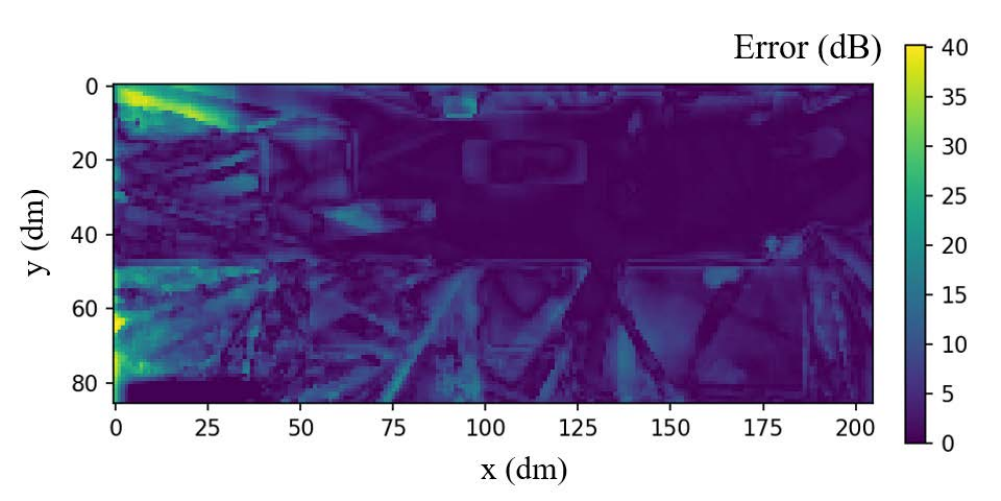}
        \vspace{-20pt}
        \caption{}
        \label{fig:test_err_0_6}
    \end{subfigure}
    \vspace{-1.9pt}
    \caption{Comparison of the simulated path-loss heatmap obtained with Sionna RT (a) and the predicted map (b) for a test sample at a receiver height of 0.6 m. Error map (c) shows the absolute error between (a) and (b).}
    \label{fig:test_0.6}
\end{figure}

\begin{figure}[t]
    \centering
    \begin{subfigure}[b]{0.4\textwidth}
        \centering
        \includegraphics[width=\textwidth]{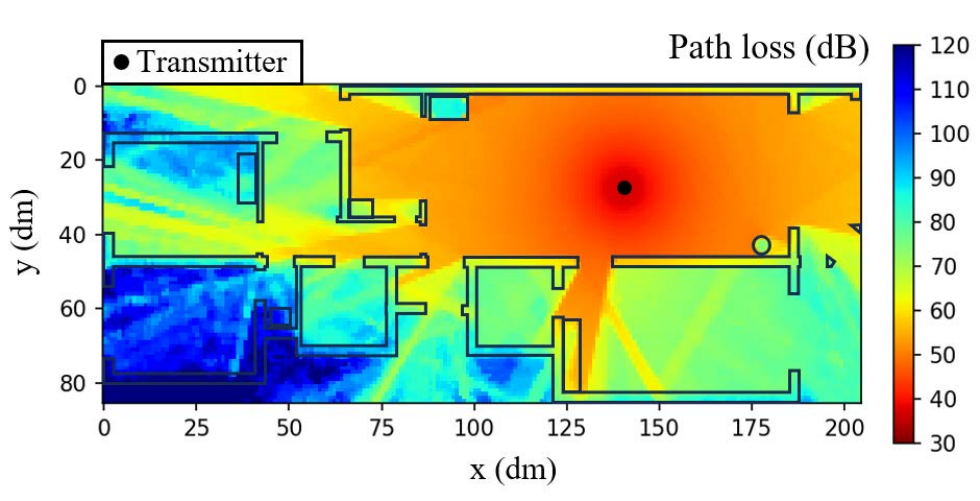}
        \vspace{-20pt}
        \caption{}
        \label{fig:test_gt_1_5}
    \end{subfigure}
    \vspace{-1.9pt}
    \vfill
    \begin{subfigure}[b]{0.4\textwidth}
        \centering
        \includegraphics[width=\textwidth]{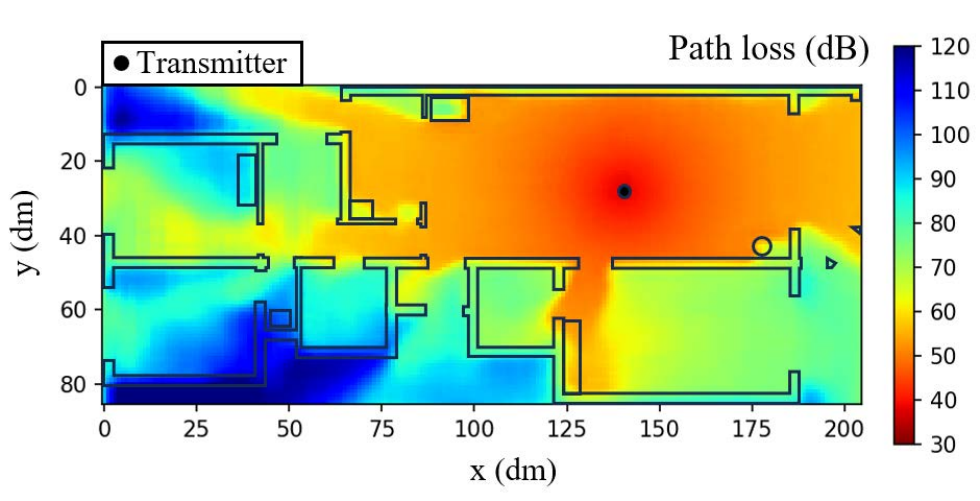}
        \vspace{-20pt}
        \caption{}
        \label{fig:test_pred_1_5}
    \end{subfigure}
    \vspace{-1.9pt}
    \vfill
    \begin{subfigure}[b]{0.4\textwidth}
        \centering
        \includegraphics[width=\textwidth]{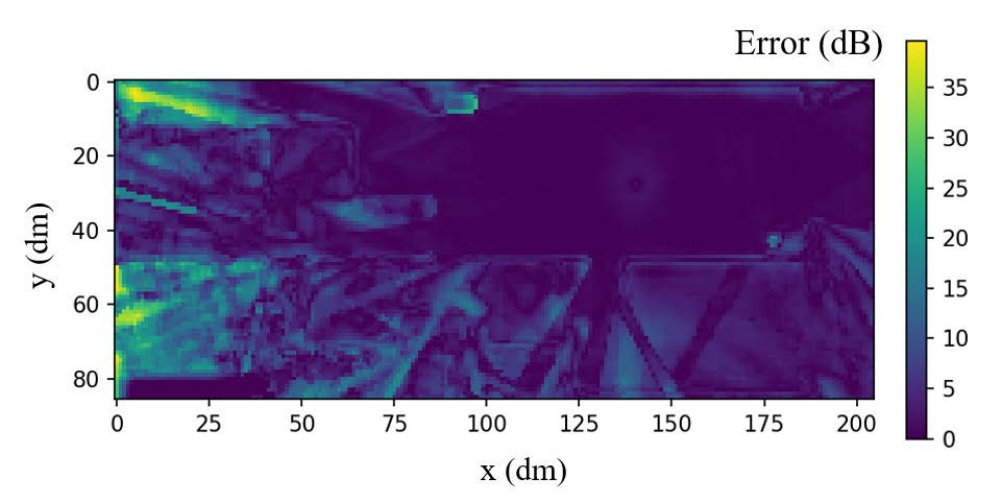}
        \vspace{-20pt}
        \caption{}
        \label{fig:test_err_1_5}
    \end{subfigure}
    \vspace{-1.9pt}
    \caption{Comparison of the simulated path-loss heatmap obtained with Sionna RT (a) and the predicted map (b) for a test sample at a receiver height of 1.5 m. Error map (c) shows the absolute error between (a) and (b).}
    \label{fig:test_1.5}
\end{figure}

\section{Evaluation}
This section presents the experimental evaluation of the proposed SenseRay-3D framework. The experiments aim to validate its prediction accuracy, generalization capability, and inference efficiency under realistic indoor propagation scenarios. The evaluation is conducted on the synthetic dataset derived in section~V. The following subsections describe the experimental setup, performance metrics, and quantitative results, followed by visual comparisons and error analysis at different receiver heights.

\begin{figure}[t]
    \centering
    \begin{subfigure}[b]{0.4\textwidth}
        \centering
        \includegraphics[width=\textwidth]{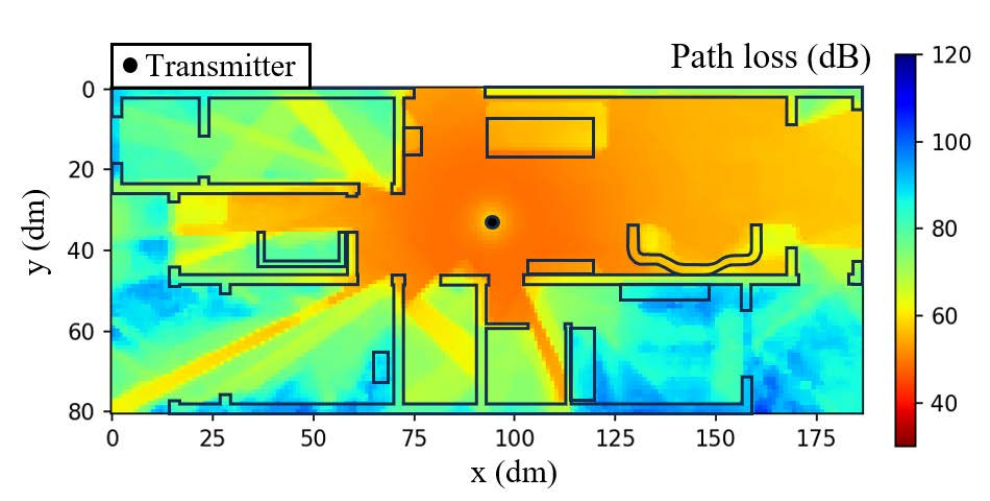}
        \vspace{-20pt}
        \caption{}
        \label{fig:val_gt_0_6}
    \end{subfigure}
    \vspace{-1.9pt}
    \vfill
    \begin{subfigure}[b]{0.4\textwidth}
        \centering
        \includegraphics[width=\textwidth]{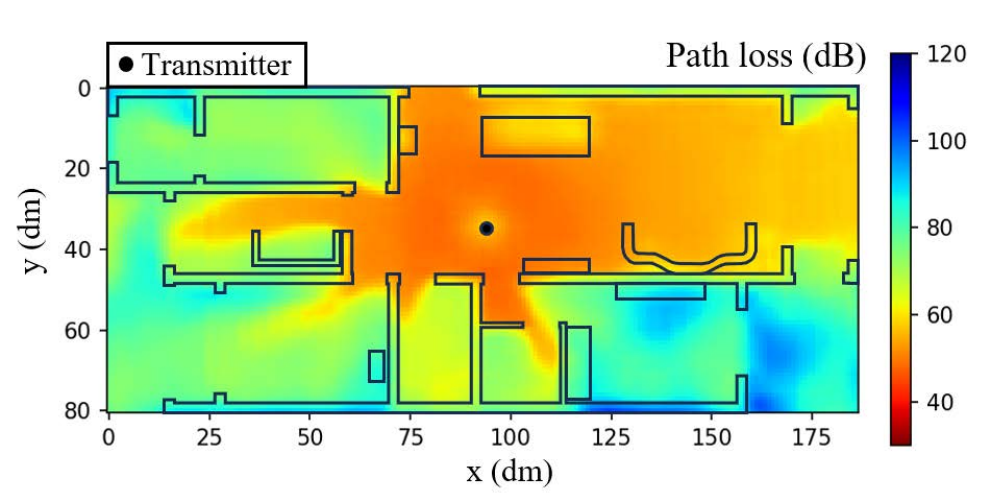}
        \vspace{-20pt}
        \caption{}
        \label{fig:val_pred_0_6}
    \end{subfigure}
    \vspace{-1.9pt}
    \vfill
    \begin{subfigure}[b]{0.4\textwidth}
        \centering
        \includegraphics[width=\textwidth]{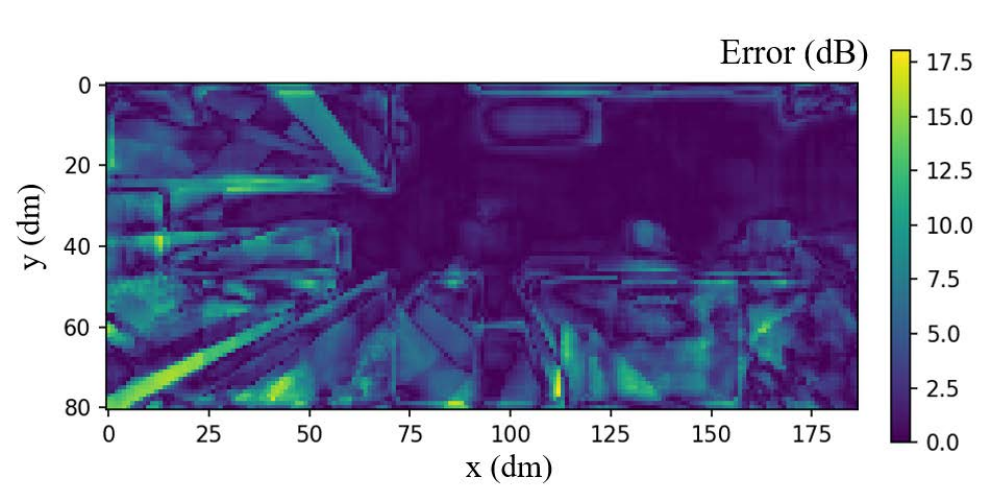}
        \vspace{-20pt}
        \caption{}
        \label{fig:val_err_0_6}
    \end{subfigure}
    \vspace{-1.9pt}
    \caption{Comparison of the simulated path-loss heatmap obtained with Sionna RT (a) and the predicted map (b) for a validation sample at a receiver height of 0.6 m. Error map (c) shows the absolute error between (a) and (b).}
    \label{fig:val_0.6}
\end{figure}

\begin{figure}[t]
    \centering
    \begin{subfigure}[b]{0.4\textwidth}
        \centering
        \includegraphics[width=\textwidth]{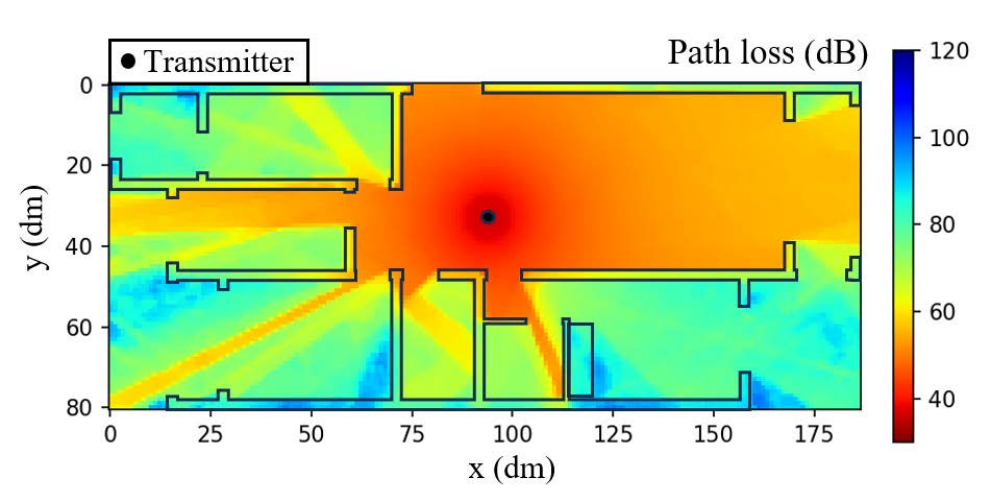}
        \vspace{-20pt}
        \caption{}
        \label{fig:val_gt_1_5}
    \end{subfigure}
    \vspace{-1.9pt}
    \vfill
    \begin{subfigure}[b]{0.4\textwidth}
        \centering
        \includegraphics[width=\textwidth]{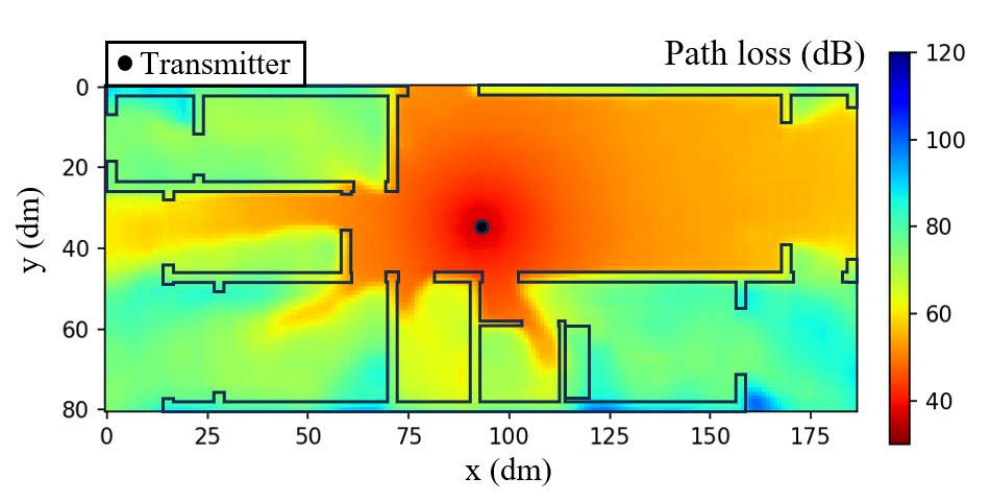}
        \vspace{-20pt}
        \caption{}
        \label{fig:val_pred_1_5}
    \end{subfigure}
    \vspace{-1.9pt}
    \vfill
    \begin{subfigure}[b]{0.4\textwidth}
        \centering
        \includegraphics[width=\textwidth]{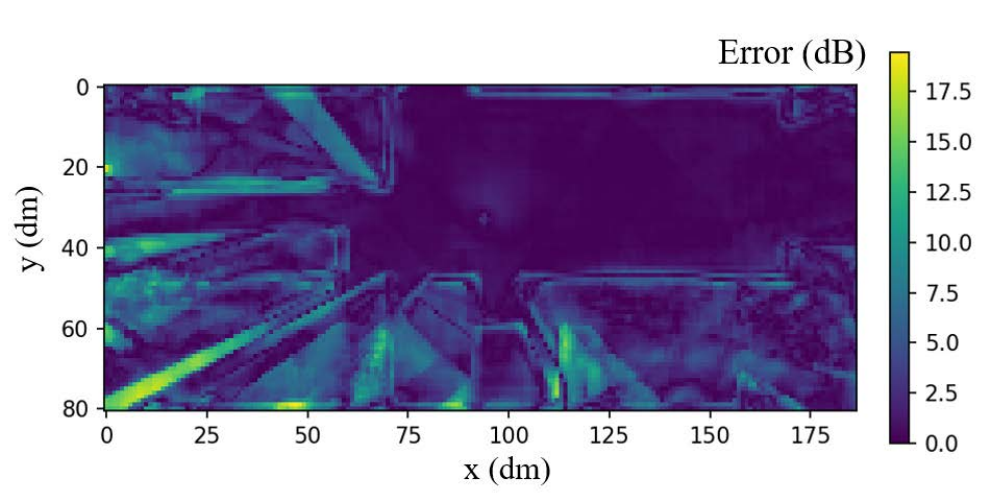}
        \vspace{-20pt}
        \caption{}
        \label{fig:val_err_1_5}
    \end{subfigure}
    \vspace{-1.9pt}

    \caption{Comparison of the simulated path-loss heatmap obtained with Sionna RT (a) and the predicted map (b) for a validation sample at a receiver height of 1.5 m. Error map (c) shows the absolute error between (a) and (b).}
    \label{fig:val_1.5}
\end{figure}

\subsection{Experimental Setup}

A total of 1805 samples are produced from the 56 apartments. The dataset is split at the scene level to avoid spatial leakage, ensuring that transmitter positions from the same environment do not appear across different subsets. To increase data diversity and improve model generalization, rotational augmentation is performed by rotating each scene around the XY-plane by 90°, 180°, and 270°, thereby tripling the effective training set size without additional simulations. Specifically, 80\% of the samples are used for training and 20\% for validation.

For evaluation, six entirely unseen indoor environments are used as the test set. This configuration aims to assess the generalization capability of the proposed model to new apartment geometries and transmitter coordinates not present in the training or validation data. Specifically, the test set is designed to evaluate two aspects: 
1) the model’s ability to generalize to previously unseen layouts with different sizes and furniture configurations; and 2) its capability to predict path-loss heatmaps for transmitter positions not encountered during training.
Table~\ref{tab:split} summarizes the distinction among the different dataset splits. The terms “known” and “unknown” indicate whether the geometry or transmitter positions are included in the training dataset. Training is conducted on known geometries with known transmitter positions. The validation set uses the same geometries as the training set but features unseen transmitter coordinates, whereas the test set comprises completely unseen geometries with unseen transmitter coordinates.

\begin{table}[t]
\centering
\renewcommand{\arraystretch}{1.2}
\setlength{\tabcolsep}{10pt}
\begin{tabular}{|c|c|c|}
\hline
 & \textbf{Transmitter Position} & \textbf{Geometry} \\ \hline
Training   & Known   & Known   \\ \hline
Validation & Unknown & Known   \\ \hline
Test       & Unknown & Unknown \\ \hline
\end{tabular}
\caption{Indicating whether the 
transmitter positions and scene geometries are known or unknown 
with respect to the training set.}
\label{tab:split}
\end{table}
Training is implemented in PyTorch on a single NVIDIA H100 GPU with 80 GB memory.
Mini-batches of four samples are randomly drawn from the training set using PyTorch’s DataLoader with shuffling enabled, which ensures diversity of transmitter locations and scene geometries across iterations. The model is trained for 300 epochs using the AdamW optimizer with an initial learning rate of $1\times 10^{-4}$ and a weight decay of $1\times 10^{-2}$. The learning rate follows a cosine-annealing schedule with warm restarts, and automatic mixed precision is enabled to reduce memory consumption and accelerate training. Early stopping based on the validation loss is applied to prevent overfitting. All input features are normalized using scene-level statistics. During training, a masked root mean squared error (RMSE) loss is employed, which computes the square root of the mean squared difference between the predicted and ground-truth path-loss residuals over valid voxels only.

\subsection{Evaluation Metrics}

To quantitatively assess the accuracy of the predicted path-loss heatmaps, two error metrics are employed: the mean absolute error (MAE) and RMSE. For each voxel $v$, let $\hat{P}(t,v)$ and $P(t,v)$ denote the predicted and ground-truth received-power values in dB, respectively. The MAE is defined as
\begin{equation}
    \text{MAE} = \frac{1}{N} \sum_{v=1}^{N} 
    \big| \hat{P}(t,v) - P(t,v) \big|,
\end{equation}
and the RMSE is given by
\begin{equation}
    \text{RMSE} = \sqrt{ \frac{1}{N} \sum_{v=1}^{N} 
    \big( \hat{P}(t,v) - P(t,v) \big)^2 } ,
\end{equation}
where $N$ denotes the number of valid voxels in the path-loss heatmaps.  

The MAE measures the average prediction deviation in dB, while the RMSE penalizes larger errors and is therefore more sensitive to abrupt variations caused by shadowing or strong reflections. Together, these metrics provide a comprehensive evaluation of the model’s prediction accuracy.

\subsection{Evaluation Result}

\begin{figure}[]
    \centering
    \begin{subfigure}[b]{0.35\textwidth}
        \centering
        \includegraphics[width=\textwidth]{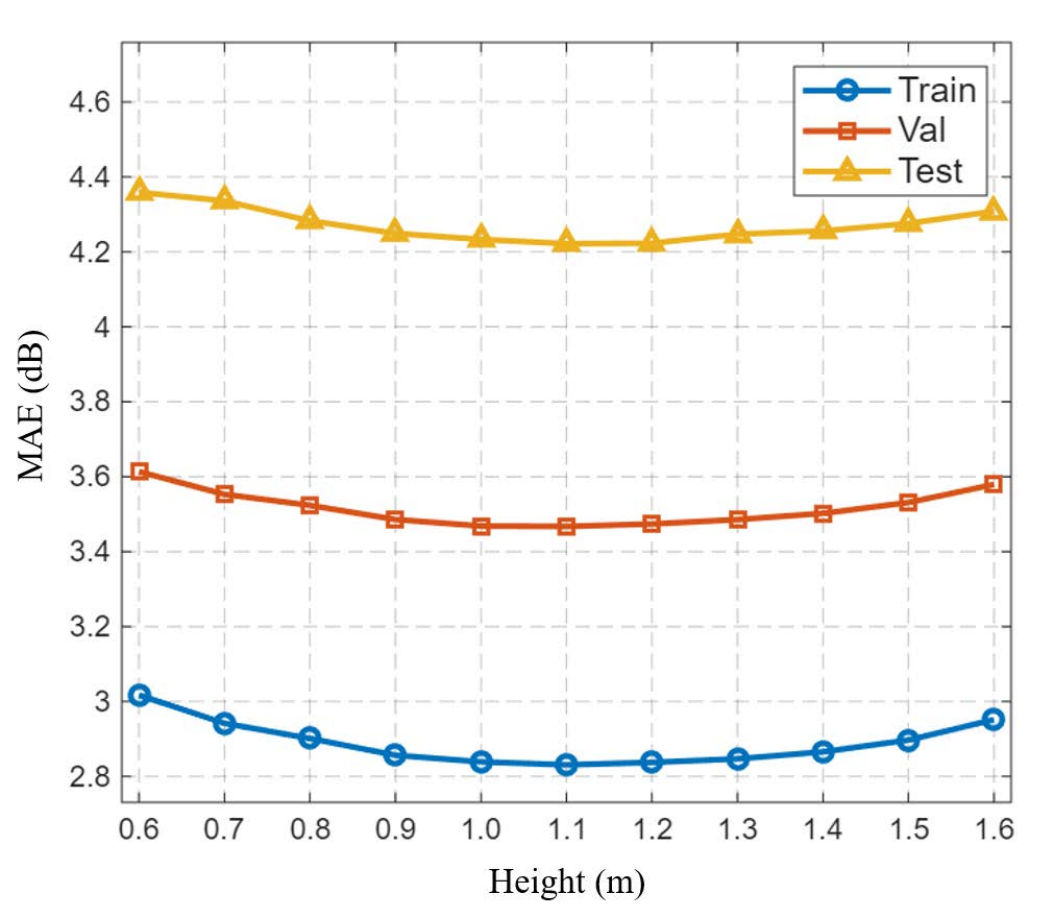}
        \vspace{-15pt}
        \caption{}
        \label{fig:test_gt}
    \end{subfigure}
    \hfill
    \begin{subfigure}[b]{0.35\textwidth}
        \centering
        \includegraphics[width=\textwidth]{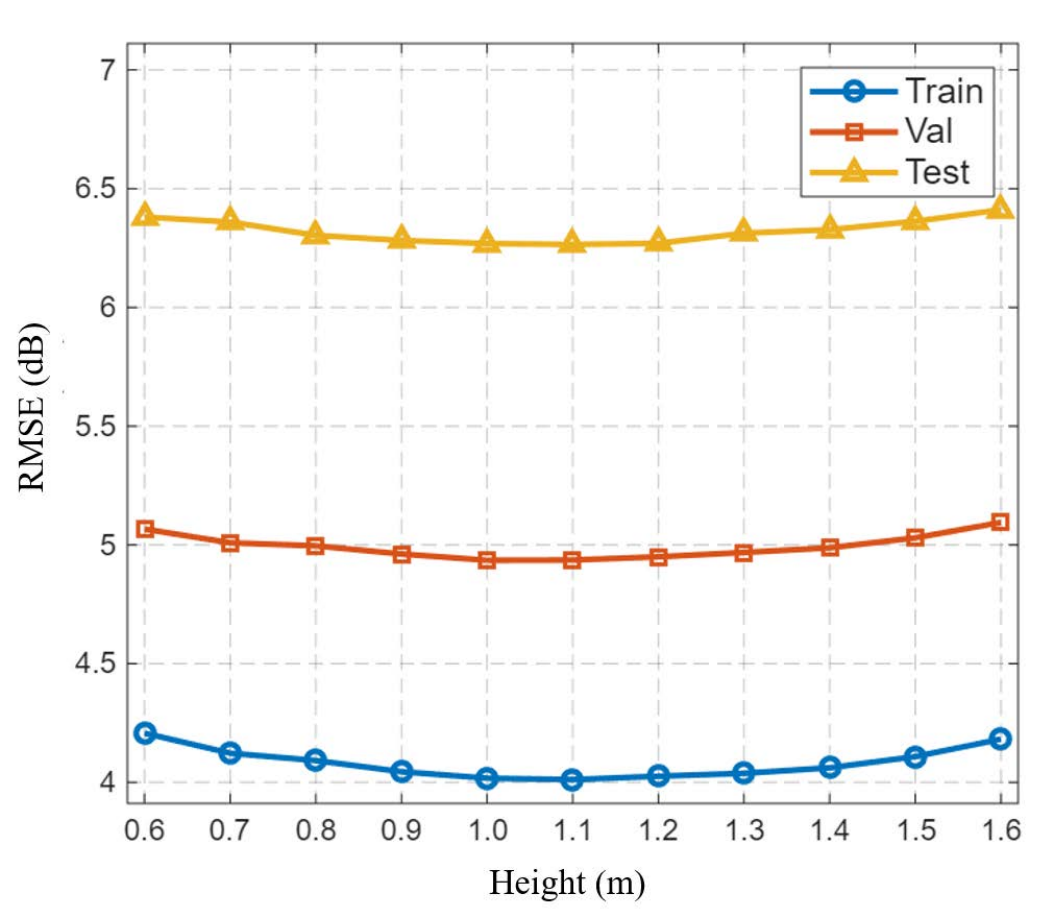}
        \vspace{-15pt}
        \caption{}
        \label{fig:test_err}
    \end{subfigure}
    \caption{(a) MAE and (b) RMSE at different receiver heights for 
training, validation, and test sets.}
    \label{fig:MAE_RMSE}
\end{figure}

\begin{table}[t]
\centering
\renewcommand{\arraystretch}{1.2}
\setlength{\tabcolsep}{10pt}
\begin{tabular}{|c|c|c|c|}
\hline
 & \textbf{MAE(dB)} & \textbf{RMSE(dB)} & \textbf{Samples} \\ \hline
Training & 2.88 & 4.08 & 1446 \\ \hline
Validation & 3.51 & 4.99 & 359 \\ \hline
Test & 4.27 & 6.32 & 225 \\ \hline
\end{tabular}
\caption{Quantitative evaluation results on training, validation, and test sets.}
\label{tab:result}
\end{table}
The overall quantitative results for all dataset splits are summarized in Table~\ref{tab:result}. On the training set, SenseRay-3D achieves an MAE of 2.88~dB and an RMSE of 4.08~dB, demonstrating high fidelity under known Tx positions and geometries. On the validation set, where Tx positions are unseen but geometries are known, the errors increase slightly to 3.51~dB MAE and 4.99~dB RMSE, indicating strong generalization to new Tx coordinates. The most challenging scenario is the test set, which includes both unseen Tx positions and geometries. Even under these conditions, the model maintains reasonable accuracy, achieving an MAE of 4.27~dB and an RMSE of 6.32~dB across 225 samples. These results confirm that the proposed method generalizes effectively from known to unknown environments while preserving robust prediction accuracy.

Figs.~\ref{fig:val_0.6}–\ref{fig:test_1.5} present qualitative examples from the validation and test sets at receiver heights of 0.6~m and 1.5~m, respectively. In each figure, (a) shows the simulated ground truth obtained from Sionna RT, (b) depicts the predicted path-loss heatmap generated by the SwinUNETR model, and (c) illustrates the absolute error in dB. Since the model outputs only the path-loss heatmaps, the corresponding cross-sectional layout at the target receiver height is overlaid to better visualize the scene geometry.

Across all cases, the predicted path-loss heatmaps closely match the ground truth, with the strongest agreement observed in line-of-sight regions near the Tx. The model captures large-scale attenuation trends as well as fine-grained effects such as through-wall penetration and shadowing caused by furniture and partitions. The error maps indicate that most discrepancies occur along material boundaries and in distant NLoS regions, where signal power is lower and multipath components dominate. This behavior is consistent with expectations, as prediction errors typically increase in low-power areas with complex propagation conditions. Importantly, these regions have limited impact on system-level performance, since users located far from the transmitter are usually served by other access points in practical deployments.

Figure~\ref{fig:MAE_RMSE} further illustrates the variation of MAE and RMSE across receiver heights from 0.6~m to 1.6~m. The results show stable prediction accuracy across all heights, with minimal fluctuations. Errors are lowest around 1.1~m and increase slightly near the floor and ceiling (0.6~m and 1.6~m), which can be attributed to stronger multipath effects introducing additional propagation complexity. The consistent separation among the training, validation, and test curves demonstrates the robust generalization capability of SenseRay-3D across spatial configurations.

Finally, the model achieves an average inference time of 217~ms per sample on an NVIDIA RTX~5080 GPU, confirming that the proposed framework can support near real-time path-loss prediction for large-scale indoor environments.

\section{Conclusion}

This paper presented SenseRay-3D, a generalizable and physics-informed framework for end-to-end indoor propagation modeling. By integrating RGB-D sensing, voxelized physics-aware representations, and a SwinUNETR-based neural predictor, SenseRay-3D directly infers path-loss heatmaps from perceptual scene data without requiring manual geometry reconstruction or material annotation, thereby extending our previous EM DeepRay framework from 2D layer-wise estimation to a unified end-to-end 3D modeling paradigm. 
A large-scale synthetic dataset was also constructed to support validation. Experimental results demonstrated that SenseRay-3D achieves an MAE of 4.27 dB on completely unseen environments and an average inference time of 217 ms per sample, confirming its scalability, efficiency, and predictive accuracy. Further analyses showed that the model captures large-scale attenuation trends and fine-grained multipath effects, maintaining stable performance across different receiver heights. These findings underscore the potential of combining physical priors with perceptual scene sensing to enable data-efficient and generalizable radio propagation modeling.
As part of our previous work, EM DeepRay has been successfully integrated into the Ranplan Professional for real-world network planning. Building on this foundation, future research will explore the integration of SenseRay-3D to validate its engineering applicability and commercial potential in sense-driven network design. Moreover, the developed dataset will be publicly released to stimulate further research in indoor propagation modeling.

\balance

\bibliographystyle{ieeetr}
\bibliography{references}

@ARTICLE{Friis1946transmission,
  author={Friis, H.T.},
  journal={Proceedings of the IRE}, 
  title={A Note on a Simple Transmission Formula}, 
  year={1946},
  volume={34},
  number={5},
  pages={254-256},
  doi={10.1109/JRPROC.1946.234568}}

@inproceedings{fu20213d,
  title={3d-front: 3d furnished rooms with layouts and semantics},
  author={Fu, Huan and Cai, Bowen and Gao, Lin and Zhang, Ling-Xiao and Wang, Jiaming and Li, Cao and Zeng, Qixun and Sun, Chengyue and Jia, Rongfei and Zhao, Binqiang and others},
  booktitle={Proceedings of the IEEE/CVF International Conference on Computer Vision},
  pages={10933--10942},
  year={2021}
}

@article{Denninger2023, 
    doi = {10.21105/joss.04901},
    url = {https://doi.org/10.21105/joss.04901},
    year = {2023},
    publisher = {The Open Journal}, 
    volume = {8},
    number = {82},
    pages = {4901}, 
    author = {Maximilian Denninger and Dominik Winkelbauer and Martin Sundermeyer and Wout Boerdijk and Markus Knauer and Klaus H. Strobl and Matthias Humt and Rudolph Triebel},
    title = {BlenderProc2: A Procedural Pipeline for Photorealistic Rendering}, 
    journal = {Journal of Open Source Software}
}

@techreport{ITU_R_P2040_3_2023,
  title        = {{Effects of Building Materials and Structures on Radiowave Propagation above about 100 MHz}},
  author       = {{ITU-R}},
  institution  = {ITU-R},
  number       = {P.2040-3},
  year         = {2023},
  month        = {August},
  day          = {23},
  type         = {Recommendation},
  address      = {Geneva, Switzerland},
  pages        = {28},
  howpublished = {Available: \url{https://www.itu.int/rec/R-REC-P.2040-3-202308-I/en}},
}

@book{balanis2012advanced,
  title={Advanced engineering electromagnetics},
  author={Balanis, Constantine A},
  year={2012},
  publisher={John Wiley \& Sons}
}

@book{cheng2014field,
  title={Field and Wave Electromagnetics},
  author={Cheng, D.K.},
  isbn={9781292026565},
  series={Addison-Wesley series in electrical engineering},
  url={https://books.google.co.uk/books?id=nQFKngEACAAJ},
  year={2014},
  publisher={Pearson Education Limited}
}

@ARTICLE{9298918,
  author={Bakirtzis, Stefanos and Hashimoto, Takahiro and Sarris, Costas D.},
  journal={IEEE Transactions on Antennas and Propagation}, 
  title={{FDTD-Based Diffuse Scattering and Transmission Models for Ray Tracing of Millimeter-Wave Communication Systems}}, 
  year={2021},
  volume={69},
  number={6},
  pages={3389-3398},
  doi={10.1109/TAP.2020.3044373}}

@ARTICLE{1232163,
  author={Sarkar, T.K. and Zhong Ji and Kyungjung Kim and Medouri, A. and Salazar-Palma, M.},
  journal={IEEE Antennas and Propagation Magazine}, 
  title={A survey of various propagation models for mobile communication}, 
  year={2003},
  volume={45},
  number={3},
  pages={51-82},
  doi={10.1109/MAP.2003.1232163}}

@INPROCEEDINGS{sionnart,
  author={Hoydis, Jakob and Aoudia, Faycal Ait and Cammerer, Sebastian and Nimier-David, Merlin and Binder, Nikolaus and Marcus, Guillermo and Keller, Alexander},
  booktitle={2023 IEEE Globecom Workshops (GC Wkshps)}, 
  title={Sionna RT: Differentiable Ray Tracing for Radio Propagation Modeling}, 
  year={2023},
  volume={},
  number={},
  pages={317-321},
  doi={10.1109/GCWkshps58843.2023.10465179}}

@INPROCEEDINGS{sionna,
  author={Lee, Ju-Hyung and Kim, Joongheon},
  booktitle={2023 International Conference on Information Networking (ICOIN)}, 
  title={Sionna: Introduction to Embedded Open-Source Semantic Communication Platforms}, 
  year={2023},
  volume={},
  number={},
  pages={775-777},
  doi={10.1109/ICOIN56518.2023.10048938}}

@inproceedings{hdf5,
  title={{An overview of the {HDF5} technology suite and its applications}},
  author={Folk, Mike and Heber, Gerd and Koziol, Quincey and Pourmal, Elena and Robinson, Dana},
  booktitle={Proceedings of the EDBT/ICDT 2011 workshop on array databases},
  pages={36--47},
  year={2011}
}

@inproceedings{he2023swinunetrv2,
  title={Swinunetr-v2: Stronger swin transformers with stagewise convolutions for 3d medical image segmentation},
  author={He, Yufan and Nath, Vishwesh and Yang, Dong and Tang, Yucheng and Myronenko, Andriy and Xu, Daguang},
  booktitle={International Conference on Medical Image Computing and Computer-Assisted Intervention},
  pages={416--426},
  year={2023},
  organization={Springer}
}

@inproceedings{swinunetr,
author = {Hatamizadeh, Ali and Nath, Vishwesh and Tang, Yucheng and Yang, Dong and Roth, Holger R. and Xu, Daguang},
title = {{Swin UNETR}: Swin Transformers for Semantic Segmentation of Brain Tumors in {MRI} Images},
year = {2021},
isbn = {978-3-031-08998-5},
publisher = {Springer-Verlag},
address = {Berlin, Heidelberg},
url = {https://doi.org/10.1007/978-3-031-08999-2_22},
doi = {10.1007/978-3-031-08999-2_22},
booktitle = {Brainlesion: Glioma, Multiple Sclerosis, Stroke and Traumatic Brain Injuries: 7th International Workshop, BrainLes 2021, Held in Conjunction with MICCAI 2021, Virtual Event, September 27, 2021, Revised Selected Papers, Part I},
pages = {272–284},
numpages = {13}
}

@InProceedings{UNETR,
  author    = {Hatamizadeh, Ali and Tang, Yucheng and Nath, Vishwesh and Yang, Dong and Myronenko, Andriy and Landman, Bennett and Roth, Holger R. and Xu, Daguang},
  title     = {{UNETR}: Transformers for {3D} Medical Image Segmentation},
  booktitle = {Proceedings of the IEEE/CVF Winter Conference on Applications of Computer Vision (WACV)},
  month     = {January},
  year      = {2022},
  pages     = {574-584}
}

@InProceedings{swintransmformer,
  author    = {Liu, Ze and Lin, Yutong and Cao, Yue and Hu, Han and Wei, Yixuan and Zhang, Zheng and Lin, Stephen and Guo, Baining},
  title     = {Swin Transformer: Hierarchical Vision Transformer Using Shifted Windows},
  booktitle = {Proceedings of the IEEE/CVF International Conference on Computer Vision (ICCV)},
  month     = {October},
  year      = {2021},
  pages     = {10012-10022}
}

@InProceedings{Tang_2022_CVPR,
  author    = {Tang, Yucheng and Yang, Dong and Li, Wenqi and Roth, Holger R. and Landman, Bennett and Xu, Daguang and Nath, Vishwesh and Hatamizadeh, Ali},
  title     = {Self-Supervised Pre-Training of Swin Transformers for {3D} Medical Image Analysis},
  booktitle = {Proceedings of the IEEE/CVF Conference on Computer Vision and Pattern Recognition (CVPR)},
  month     = {June},
  year      = {2022},
  pages     = {20730-20740}
}

@ARTICLE{9040264,
  author={Giordani, Marco and Polese, Michele and Mezzavilla, Marco and Rangan, Sundeep and Zorzi, Michele},
  journal={IEEE Communications Magazine}, 
  title={Toward {6G} Networks: Use Cases and Technologies}, 
  year={2020},
  volume={58},
  number={3},
  pages={55-61},
  doi={10.1109/MCOM.001.1900411}}

@book{siomina2007radio,
  title={Radio Network Planning and Resource Optimization: Mathematical Models and Algorithms for {UMTS}, {WLANs}, and Ad Hoc Networks},
  author={Siomina, Iana},
  year={2007},
  publisher={Linkopings Universitet (Sweden)}
}

@article{wang2022reflection,
  title={Reflection characteristics measurements of indoor wireless link in {D-band}},
  author={Wang, Mingxu and Wang, Yanyi and Li, Weiping and Ding, Junjie and Bian, Chengzhen and Wang, Xinyi and Wang, Chao and Li, Chao and Zhong, Zhimeng and Yu, Jianjun},
  journal={Sensors},
  volume={22},
  number={18},
  pages={6908},
  year={2022},
  publisher={MDPI}
}

@article{obeidat2020indoor,
  title={Indoor environment propagation review},
  author={Obeidat, H and Alabdullah, A and Elkhazmi, E and Suhaib, W and Obeidat, Omar and Alkhambashi, Majid and Mosleh, M and Ali, N and Dama, Y and Abidin, Z and others},
  journal={Computer Science Review},
  volume={37},
  pages={100272},
  year={2020},
  publisher={Elsevier}
}

@INPROCEEDINGS{6927716,
  author={Alhamoud, Alaa and Kreger, Michael and Afifi, Haitham and Gottron, Christian and Burgstahler, Daniel and Englert, Frank and Böhnstedt, Doreen and Steinmetz, Ralf},
  booktitle={39th Annual IEEE Conference on Local Computer Networks Workshops}, 
  title={Empirical investigation of the effect of the door's state on received signal strength in indoor environments at 2.4 {GHz}}, 
  year={2014},
  volume={},
  number={},
  pages={652-657},
  doi={10.1109/LCNW.2014.6927716}}

@INPROCEEDINGS{1543252,
  author={Abhayawardhana, V.S. and Wassell, I.J. and Crosby, D. and Sellars, M.P. and Brown, M.G.},
  booktitle={2005 IEEE 61st Vehicular Technology Conference}, 
  title={Comparison of empirical propagation path loss models for fixed wireless access systems}, 
  year={2005},
  volume={1},
  number={},
  pages={73-77 Vol. 1},
  doi={10.1109/VETECS.2005.1543252}}

@ARTICLE{10599118,
  author={Zhu, Ethan and Sun, Haijian and Ji, Mingyue},
  journal={IEEE Wireless Communications}, 
  title={Physics-Informed Generalizable Wireless Channel Modeling with Segmentation and Deep Learning: Fundamentals, Methodologies, and Challenges}, 
  year={2024},
  volume={31},
  number={6},
  pages={170-177},
  doi={10.1109/MWC.015.2300603}}

@INPROCEEDINGS{10188258,
  author={Bhatia, Gurjot Singh and Corre, Yoann and Di Renzo, M.},
  booktitle={2023 Joint European Conference on Networks and Communications \& 6G Summit (EuCNC/6G Summit)}, 
  title={Efficient Ray-Tracing Channel Emulation in Industrial Environments: An Analysis of Propagation Model Impact}, 
  year={2023},
  volume={},
  number={},
  pages={180-185},
  doi={10.1109/EuCNC/6GSummit58263.2023.10188258}}

@ARTICLE{9496115,
  author={Seretis, Aristeidis and Sarris, Costas D.},
  journal={IEEE Transactions on Antennas and Propagation}, 
  title={An Overview of Machine Learning Techniques for Radiowave Propagation Modeling}, 
  year={2022},
  volume={70},
  number={6},
  pages={3970-3985},
  doi={10.1109/TAP.2021.3098616}}

@ARTICLE{8743390,
  author={Sun, Yaohua and Peng, Mugen and Zhou, Yangcheng and Huang, Yuzhe and Mao, Shiwen},
  journal={IEEE Communications Surveys \& Tutorials}, 
  title={Application of Machine Learning in Wireless Networks: Key Techniques and Open Issues}, 
  year={2019},
  volume={21},
  number={4},
  pages={3072-3108},
  doi={10.1109/COMST.2019.2924243}}

@ARTICLE{10640063,
  author={Vasudevan, Manjuladevi and Yuksel, Murat},
  journal={IEEE Open Journal of the Communications Society}, 
  title={Machine Learning for Radio Propagation Modeling: A Comprehensive Survey}, 
  year={2024},
  volume={5},
  number={},
  pages={5123-5153},
  doi={10.1109/OJCOMS.2024.3446457}}

@ARTICLE{10812728,
  author={Wen, Dingzhu and Zhou, Yong and Li, Xiaoyang and Shi, Yuanming and Huang, Kaibin and Letaief, Khaled B.},
  journal={IEEE Communications Surveys \& Tutorials}, 
  title={A Survey on Integrated Sensing, Communication, and Computation}, 
  year={2024},
  volume={},
  number={},
  pages={1-1},
  doi={10.1109/COMST.2024.3521498}}

@ARTICLE{10430216,
  author={Zeng, Yong and Chen, Junting and Xu, Jie and Wu, Di and Xu, Xiaoli and Jin, Shi and Gao, Xiqi and Gesbert, David and Cui, Shuguang and Zhang, Rui},
  journal={IEEE Communications Surveys \& Tutorials}, 
  title={A Tutorial on Environment-Aware Communications via Channel Knowledge Map for {6G}}, 
  year={2024},
  volume={26},
  number={3},
  pages={1478-1519},
  doi={10.1109/COMST.2024.3364508}}

@ARTICLE{10949588,
  author={Bai, Lu and Huang, Ziwei and Sun, Mingran and Cheng, Xiang and Cui, Lizhen},
  journal={IEEE Communications Surveys \& Tutorials}, 
  title={Multi-Modal Intelligent Channel Modeling: A New Modeling Paradigm via Synesthesia of Machines}, 
  year={2025},
  volume={},
  number={},
  pages={1-1},
  doi={10.1109/COMST.2025.3558046}}

@incollection{mogensen1999cost,
  title={COST Action 231: Digital Mobile Radio Towards Future Generation System, Final Report.},
  author={Mogensen, Preben Elgaard and Wigard, Jeroen},
  booktitle={Section 5.2: On antenna and frequency diversity in {GSM}. Section 5.3: Capacity study of frequency hopping {GSM} network},
  year={1999}
}

@techreport{ITU-RP1238-12,
  author      = {{ITU-R}},
  title       = {Propagation data and prediction methods for the planning of indoor radiocommunication systems and radio local area networks in the frequency range 300 {MHz} to 450 {GHz}},
  institution = {International Telecommunication Union},
  type        = {Recommendation ITU-R P.1238-12},
  year        = {2023},
  address     = {Geneva, Switzerland},
  month       = {aug},
  url         = {https://www.itu.int/rec/R-REC-P.1238}
}

@ARTICLE{103807,
  author={McKown, J.W. and Hamilton, R.L.},
  journal={IEEE Network}, 
  title={Ray tracing as a design tool for radio networks}, 
  year={1991},
  volume={5},
  number={6},
  pages={27-30},
  doi={10.1109/65.103807}}

@INPROCEEDINGS{10464960,
  author={Yu, Gang and Zhou, Lingyou and Zhang, Jiliang and Zhang, Jie},
  booktitle={2023 IEEE Globecom Workshops (GC Wkshps)}, 
  title={A Ray-Launching Algorithm for Polarized Wireless Channel Prediction}, 
  year={2023},
  volume={},
  number={},
  pages={1928-1933},
  doi={10.1109/GCWkshps58843.2023.10464960}}

@ARTICLE{5979136,
  author={Austin, Andrew C. M. and Neve, Michael J. and Rowe, Gerard B.},
  journal={IEEE Transactions on Antennas and Propagation}, 
  title={Modeling Propagation in Multifloor Buildings Using the {FDTD} Method}, 
  year={2011},
  volume={59},
  number={11},
  pages={4239-4246},
  doi={10.1109/TAP.2011.2164181}}

@ARTICLE{6748900,
  author={Azpilicueta, Leire and Rawat, Meenakshi and Rawat, Karun and Ghannouchi, Fadhel M. and Falcone, Francisco},
  journal={IEEE Transactions on Antennas and Propagation}, 
  title={A Ray Launching-Neural Network Approach for Radio Wave Propagation Analysis in Complex Indoor Environments}, 
  year={2014},
  volume={62},
  number={5},
  pages={2777-2786},
  doi={10.1109/TAP.2014.2308518}}

@INPROCEEDINGS{1696368,
  author={Popescu, I. and Nikitopoulos, D. and Nafornita, I. and Constantinou, P.},
  booktitle={2006 IEEE International Conference on Wireless and Mobile Computing, Networking and Communications}, 
  title={{ANN} Prediction Models for Indoor Environment}, 
  year={2006},
  volume={},
  number={},
  pages={366-371},
  doi={10.1109/WIMOB.2006.1696368}}

@ARTICLE{9670666,
  author={Seretis, Aristeidis and Sarris, Costas D.},
  journal={IEEE Transactions on Antennas and Propagation}, 
  title={Toward Physics-Based Generalizable Convolutional Neural Network Models for Indoor Propagation}, 
  year={2022},
  volume={70},
  number={6},
  pages={4112-4126},
  doi={10.1109/TAP.2021.3138535}}

@ARTICLE{9354618,
  author={Sotiroudis, Sotirios P. and Sarigiannidis, Panagiotis and Goudos, Sotirios K. and Siakavara, Katherine},
  journal={IEEE Access}, 
  title={Fusing Diverse Input Modalities for Path Loss Prediction: A Deep Learning Approach}, 
  year={2021},
  volume={9},
  number={},
  pages={30441-30451},
  doi={10.1109/ACCESS.2021.3059589}}

@ARTICLE{9771088,
  author={Bakirtzis, Stefanos and Chen, Jiming and Qiu, Kehai and Zhang, Jie and Wassell, Ian},
  journal={IEEE Transactions on Antennas and Propagation}, 
  title={{EM DeepRay}: An Expedient, Generalizable, and Realistic Data-Driven Indoor Propagation Model}, 
  year={2022},
  volume={70},
  number={6},
  pages={4140-4154},
  doi={10.1109/TAP.2022.3172221}}

@ARTICLE{10064304,
  author={Suga, Norisato and Maeda, Yoshihiro and Sato, Koya},
  journal={IEEE Access}, 
  title={Indoor Radio Map Construction via Ray Tracing With {RGB-D} Sensor-Based {3D} Reconstruction: Concept and Experiments in WLAN Systems}, 
  year={2023},
  volume={11},
  number={},
  pages={24863-24874},
  doi={10.1109/ACCESS.2023.3254912}}

@INPROCEEDINGS{9977926,
  author={Okamura, Wataru and Sugiyama, Kento and Ching, Gilbert Siy and Kishiki, Yukiko and Saito, Kentaro and Takada, Jun-ichi},
  booktitle={2022 IEEE 33rd Annual International Symposium on Personal, Indoor and Mobile Radio Communications (PIMRC)}, 
  title={Indoor Model Reconstruction using {3D} Point Cloud Data for Ray Tracing Simulation}, 
  year={2022},
  volume={},
  number={},
  pages={1-5},
  doi={10.1109/PIMRC54779.2022.9977926}}

@INPROCEEDINGS{10757803,
  author={Li, Yutao and Wang, Yancheng and Huang, Chuan},
  booktitle={2024 IEEE 100th Vehicular Technology Conference (VTC2024-Fall)}, 
  title={{NeRA}: Neural Reflectance and Attenuation Fields for Radio Map Reconstruction}, 
  year={2024},
  volume={},
  number={},
  pages={1-5},
  doi={10.1109/VTC2024-Fall63153.2024.10757803}}

@book{Hartley_Zisserman_2004, place={Cambridge}, edition={2}, title={Multiple View Geometry in Computer Vision}, publisher={Cambridge University Press}, author={Hartley, Richard and Zisserman, Andrew}, year={2004}}

@book{szeliski2022computer,
  title={Computer vision: algorithms and applications},
  author={Szeliski, Richard},
  year={2022},
  publisher={Springer Nature}
}

@article{zhou2018open3d,
  title={{Open3D}: A modern library for {3D} data processing},
  author={Zhou, Qian-Yi and Park, Jaesik and Koltun, Vladlen},
  journal={arXiv preprint arXiv:1801.09847},
  year={2018}
}

@ARTICLE{10949606,
  author={Suga, Norisato and Yoshida, Naoya and Gozono, Ryotaro and Maeda, Yoshihiro and Sato, Koya},
  journal={IEEE Wireless Communications Letters}, 
  title={{RGB-D} Sensor-Aided Radio Map Estimation Using Materials Classification}, 
  year={2025},
  volume={14},
  number={6},
  pages={1811-1815},
  doi={10.1109/LWC.2025.3557794}}

@article{clark2022propem,
  title= {PropEM-L: Radio Propagation Environment Modeling and Learning for Communication-Aware Multi-Robot Exploration},
  author= {Lillian Clark and Jeffrey A. Edlund and Marc Sanchez Net and Tiago Stegun Vaquero and Ali{-}akbar Agha{-}mohammadi},
  year={2022},
  month = {June},
  journal= {Robotics: Science and Systems (RSS)},
  address = {New York City, NY, USA},
  url= {https://www.roboticsproceedings.org/rss18/p014.pdf},
  clearance= {CL\#22-2105  URS305889},
  project= {subt}
}

@techreport{3GPP_TS_38_101_1_V18_8_0,
  title        = {{NR; User Equipment (UE) radio transmission and reception; Part 1: Range 1 Standalone}},
  author       = {{3rd Generation Partnership Project (3GPP)}},
  institution  = {3GPP / ETSI},
  type         = {Technical Specification},
  number       = {TS 38.101-1},
  version      = {18.8.0},
  release      = {Release 18},
  year         = {2025},
  month        = {February},
  note         = {ETSI TS 138 101-1 V18.8.0, available at: \url{https://www.etsi.org/deliver/etsi_ts/138100_138199/13810101/18.08.00_60/ts_13810101v180800p.pdf}}
}

@inproceedings{kirillov2023segment,
  title={Segment anything},
  author={Kirillov, Alexander and Mintun, Eric and Ravi, Nikhila and Mao, Hanzi and Rolland, Chloe and Gustafson, Laura and Xiao, Tete and Whitehead, Spencer and Berg, Alexander C and Lo, Wan-Yen and others},
  booktitle={Proceedings of the IEEE/CVF international conference on computer vision},
  pages={4015--4026},
  year={2023}
}

@misc{ranplan,
  title        = {Ranplan Professional},
  author       = {{Ranplan Wireless}},  
  howpublished = {[Online]. Available: \url{https://ranplanwireless.com/}},
  year         = {2025},
}

\vfill

\end{document}